\documentclass{article}

\usepackage{arxiv}

\usepackage[utf8]{inputenc} 
\usepackage[T1]{fontenc}    
\usepackage[
    colorlinks = true,
]{hyperref}
\usepackage{url}            
\usepackage{booktabs}       
\usepackage{amsfonts}       
\usepackage{nicefrac}       
\usepackage{microtype}      
\usepackage{lipsum}         
\usepackage{graphicx}
\usepackage{doi}

\usepackage{multirow}  
\usepackage{booktabs}  

\usepackage[
  autocite     = plain,
  backend      = biber,
  doi          = true,
  isbn         = false,
  url          = false,
  giveninits   = true,
  hyperref     = true,
  maxbibnames  = 10,
  maxcitenames = 1,
  style        = numeric,
  sorting      = none,
]{biblatex}
\AtEveryBibitem{\clearfield{month}}
\AtEveryCitekey{\clearfield{month}}

\AtEveryBibitem{\clearfield{editor}}
\AtEveryBibitem{\clearfield{pages}}
\AtEveryBibitem{\clearfield{volume}}
\AtEveryBibitem{\clearfield{number}}
\AtEveryBibitem{\clearfield{note}}
\AtEveryBibitem{\clearfield{urldate}}
\AtEveryBibitem{\clearfield{url}}
\AtEveryBibitem{%
\ifentrytype{article}{
  \clearfield{issn}%
  \clearfield{editor}
  \clearfield{pages}
}{}
}

\AtBeginBibliography{%
  \renewcommand*{\finalnamedelim}{%
    \ifthenelse{\value{listcount} > 2}{%
      \addcomma
      \addspace
      \bibstring{and}%
    }{%
      \addspace
      \bibstring{and}%
    }
  }
}

\renewbibmacro{in:}{%
  \ifentrytype{article}
  {%
  }%
  {%
    \printtext{\bibstring{in}\intitlepunct}%
  }%
}

\renewbibmacro*{issue+date}{%
  \setunit{\addcomma\space}
    \iffieldundef{issue}
      {\usebibmacro{date}}
      {\printfield{issue}%
       \setunit*{\addspace}%
       \usebibmacro{date}}%
  \newunit}

\renewbibmacro*{volume+number+eid}{%
  \printfield{volume}%
  \setunit*{\addcolon}%
  \printfield{number}%
  \setunit{\addcomma\space}%
  \printfield{eid}%
}

\renewbibmacro*{publisher+location+date}{%
  \printlist{publisher}%
  \setunit*{\addcomma\space}%
  \printlist{location}%
  \setunit*{\addcomma\space}%
  \usebibmacro{date}%
  \newunit%
}

\renewbibmacro*{organization+location+date}{%
  \printlist{location}%
  \setunit*{\addcomma\space}%
  \printlist{organization}%
  \setunit*{\addcomma\space}%
  \usebibmacro{date}%
  \newunit%
}

\DeclareFieldFormat{labelnumberwidth}{#1\adddot}

\DeclareFieldFormat{doi}{%
  \mkbibacro{DOI}\addcolon\addnbspace
    \ifhyperref
      {\href{http://dx.doi.org/#1}{\nolinkurl{#1}}}
      {\nolinkurl{#1}}
}

\DeclareLanguageMapping{english}{english-mimosis}

\DeclareFieldFormat{citehyperref}{%
  \DeclareFieldAlias{bibhyperref}{noformat}%
  \bibhyperref{#1}%
}

\DeclareFieldFormat{textcitehyperref}{%
  \DeclareFieldAlias{bibhyperref}{noformat}%
  \bibhyperref{%
    #1%
    \ifbool{cbx:parens}
      {\bibcloseparen\global\boolfalse{cbx:parens}}
      {}%
    }%
}

\savebibmacro{cite}
\savebibmacro{textcite}

\renewbibmacro*{cite}{%
  \printtext[citehyperref]{%
    \restorebibmacro{cite}%
    \usebibmacro{cite}}%
}

\renewbibmacro*{textcite}{%
  \ifboolexpr{
    ( not test {\iffieldundef{prenote}} and
      test {\ifnumequal{\value{citecount}}{1}} )
    or
    ( not test {\iffieldundef{postnote}} and
      test {\ifnumequal{\value{citecount}}{\value{citetotal}}} )
  }%
  {\DeclareFieldAlias{textcitehyperref}{noformat}}
  {}%
  \printtext[textcitehyperref]{%
    \restorebibmacro{textcite}%
    \usebibmacro{textcite}}%
}

\usepackage{wrapfig}

\usepackage[font={small, it}]{caption}

\usepackage[ruled, vlined, linesnumbered]{algorithm2e}

\usepackage{xcolor}

\SetCommentSty{mycommfont}
\SetArgSty{textnormal}
\newcommand{\nosemic}{\renewcommand{\@endalgocfline}{\relax}}
\newcommand{\dosemic}{\renewcommand{\@endalgocfline}{\algocf@endline}}
\let\oldnl\nl
\newcommand{\nonl}{\renewcommand{\nl}{\let\nl\oldnl}}

\usepackage{appendix}

\usepackage{mathtools}

\usepackage{mathbbol}

\usepackage[scr = dutchcal]{mathalpha}

\usepackage{amssymb}

\usepackage{blkarray}

\usepackage{multicol}

\usepackage{siunitx}

\usepackage{siunitx}

\usepackage[version=4]{mhchem}

\usepackage{listings}
\definecolor{mygray}{gray}{0.9}
\newcommand{\code}[1]{\colorbox{mygray}{\lstinline|#1|}}

\DeclareMathOperator*{\argmin}          {arg\,min}
\DeclareMathOperator*{\argmax}          {arg\,max}

\DeclareMathOperator {\abs}             {abs}

\DeclareMathOperator {\diag}            {diag}
\DeclareMathOperator {\vol}             {vol}

\DeclareMathOperator {\atan2}           {atan2}

\DeclareMathOperator {\ilr}             {ilr}
\DeclareMathOperator {\cyl}             {cyl}
\DeclareMathOperator {\mec}             {mec}

\DeclareMathOperator {\proj}            {proj}
\DeclareMathOperator {\stdz}            {stdz}
\DeclareMathSymbol{\shortminus}{\mathbin}{AMSa}{"39}

\DeclarePairedDelimiter\floor{\lfloor}{\rfloor}

\title{Flows on convex polytopes}

\author{ 
    \hspace{1mm}Tomek Diederen\\
    Institute of Molecular Systems Biology\\
    ETH Zurich\\
    \texttt{diederent@gmail.com} \\
\And
    \hspace{1mm}Nicola Zamboni \\
    Institute of Molecular Systems Biology\\
    ETH Zurich\\
    \texttt{zamboni@imsb.biol.ethz.ch} \\
}

\begin{document}
\maketitle

\begin{abstract}
    We present a framework for modeling complex, high-dimensional distributions on convex polytopes by leveraging recent advances in discrete and continuous normalizing flows on Riemannian manifolds.
    We show that any full-dimensional polytope is homeomorphic to a unit ball, and our approach harnesses flows defined on the ball, mapping them back to the original polytope.
    Furthermore, we introduce a strategy to construct flows when only the vertex representation of a polytope is available, employing maximum entropy barycentric coordinates and Aitchison geometry. 
    Our experiments take inspiration from applications in metabolic flux analysis and demonstrate that our methods achieve competitive density estimation, sampling accuracy, as well as fast training and inference times.
\end{abstract}

\section{Introduction}

We begin by outlining the two primary representations of a convex polytope. 
Next, we review the fundamentals of discrete and continuous normalizing flows, with a particular focus on their formulation on Riemannian manifolds. 
Finally, we introduce key concepts from Riemannian flow matching that motivate our use of a ball-homeomorphism to model distributions on polytopes.

\subsection{Convex polytopes}

For applications such as \ce{$^{13}$C} metabolic flux analysis, the target distribution is supported on a convex polytope. 
In this setting one infers biochemical reaction rates from isotopic labeling data, typically measured by mass-spectrometry \cite{wiechert_13c_2001, antoniewicz_guide_2018}. 
Every polytope considered here is convex. 
A polytope is defined by:

\begin{align}
\pmb{S}\, v & = h, \quad \pmb{S} \in \mathbb{R}^{M \times R}  \label{eq:eq_constraints}
\\[1mm]
\pmb{A}_c\, v & \le b_c, \quad \pmb{A} \in \mathbb{R}^{C \times R} \label{eq:ineq_constraints}
\\[1mm]
\pmb{A}^\ddagger & = 
    \begin{bmatrix}
        \pmb{S}   \\
        -\pmb{S}  \\
        \pmb{A}_c 
    \end{bmatrix}, \quad 
b^\ddagger = \begin{bmatrix}
        h  \\
        -h  \\
        b_c
\end{bmatrix} \label{eq:canon_pol}\\[1mm]
\mathcal{F}^\ddagger & = \{ v^\ddagger \in  \mathbb{R}^{R} \mid \pmb{A}^\ddagger\, v \le b^\ddagger\} \label{eq:h_rep}\\[1mm]
 & =  \{ v^\ddagger \in \mathbb{R}^R \mid v^\ddagger = \pmb{V}^\ddagger\, \lambda,\ \lambda \in \Delta_{1}^{V} \} \label{eq:v_rep}
\end{align}

Equation~\eqref{eq:eq_constraints} shows the equality constraints, while Equation~\eqref{eq:ineq_constraints} establishes the inequality constraints; matrices are shown in bold-face. 
Matrix \(\pmb{V}^\ddagger \in \mathbb{R}^{K \times V + 1}\) is a matrix whose columns are the extreme points or vertices of the polytope.
\( \mathbb{\Delta}_1^V \) is the \(V\) dimensional probability simplex embedded in \( \mathbb{R}^{V+1}\), meaning that \(\lambda_i \geq 0 \ \forall i \in \{1:V+1\}\) and \(\| \lambda\|_1 = 1\).

Equations~\eqref{eq:canon_pol}, \eqref{eq:h_rep}, and \eqref{eq:v_rep} then describe, respectively, the canonical description, the half-space (H-) representation, and the vertex (V-) representation of polytope $\mathcal{F}$.
In metabolic flux analysis, finite upper and lower bounds (as in Equation~\eqref{eq:ineq_constraints}) ensure that the polytope is a closed set and, more specifically, a Riemannian manifold with a boundary.

\subsection{Discrete normalizing flows}

Normalizing flows are a flexible way of modeling complex distributions on high-dimensional data.
In most applications, we have access to data sampled from some unknown target distribution: $y \sim p(Y)$.
A normalizing flow maps samples from a base distribution, \(q_\theta^0(Y^{0})\), to a target distribution \(q^L(Y^L)\) via a diffeomorphism \(\mathscr{f}_\theta\):

\begin{align}
    y^L & = \mathscr{f}_\theta(y^{0}), \quad y^{0} \sim q_\theta^0(Y^{0}), \quad y^{0} \in \mathcal{M}^0,\quad y^L \in \mathcal{M}^L \label{eq:norm_flow_sampling}
    \\
     & \begin{cases}
    \mathscr{f}: \mathcal{M}^0 \to \mathcal{M}^L, & \mathscr{f}^{- 1}: \mathcal{M}^L \to \mathcal{M}^0
    \\
    f \in C^{\infty}(\mathcal{M}^0, \mathcal{M}^L), & f^{-1} \in C^{\infty}(\mathcal{M}^L, \mathcal{M}^0)
    \end{cases} \label{eq:diffeomorphism}
\end{align}

The super-scripts such as $0$ and $L$ for $Y$, $y$, $\mathcal{M}$ and $q$ are not exponents, but are used to indicate distinct random variables (RV), realizations, manifolds and distributions respectively. 
\(\mathcal{M}^0, \mathcal{M}^L\) are the support of the probability density of the base and target distributions and both can be thought of as some $K$-dimensional manifold.
Equation~\eqref{eq:diffeomorphism} encapsulates the requirements for \(\mathscr{f}\) to be a diffeomorphism: it must be a bijective, smooth function whose inverse is also smooth.
The sub-script $\theta$ in \(q_\theta^0\) and \(\mathscr{f}_\theta\) indicates either hyper-parameters or parameters learned from data.
The probability density of the target distribution can be obtained by applying the change of variables formula for probability density functions:

\begin{align}
    q_\theta^L(y^L) & = q_\theta^0(y^{0}) \cdot \Bigl| \pmb{J}^{\mathscr{f}_\theta}(y^{0}) \Bigr|^{-1} \label{eq:change_of_vars}
    \\[1mm]
    & = q_\theta^0(\mathscr{f}_\theta^{-1}(y^L)) \cdot \Bigl| \pmb{J}^{\mathscr{f}_\theta^{-1}}(y^L) \Bigr| \nonumber
    \\[1mm]
    \pmb{J}^{\mathscr{f}_\theta}(y^{0}) & = \begin{bmatrix}
    \frac{\partial y_1^L}{\partial y_1^{0}} & \dots & \frac{\partial y_1^L}{\partial y_K^{0}} \\[0.5mm]
    \vdots & \ddots & \vdots \\[0.5mm]
    \frac{\partial y_K^L}{\partial y_1^{0}} & \dots & \frac{\partial y_K^L}{\partial y_K^{0}}
    \end{bmatrix} \label{eq:full_tranformation_jacobian}
\end{align}

$\pmb{J}^{\mathscr{f}_\theta}$ is the Jacobian of transformation $\mathscr{f}$ and with slight abuse of notation, $| \pmb{J}^{\mathscr{f}_\theta} |$ is the absolute value of its determinant.
The parameters of distribution $q_\theta^L$ consist of the union of the parameters of the base distribution and diffeomorphism $\mathscr{f}$. 
The goal of density estimation is to fit parameters $\theta$ from data such that variational distribution $q_\theta^L(Y^L)\approx p(Y)$.
To do so, it is necessary to evaluate the density of data via Equation \eqref{eq:change_of_vars}, thus requiring efficient computation of \(\mathscr{f}^{-1}\) and its Jacobian determinant.
Conversely, efficient computation of \(\mathscr{f}\) is required to sample from the flow via Equation \eqref{eq:norm_flow_sampling}.
A common approach to ensure efficient computation of Jacobian determinants is to design $\mathscr{f}$ so that its Jacobian matrix is triangular.

For notational brevity, we omit the superscript \(i\) for a distribution \(q^i\) when its identification is clear from the context (i.e., via the random variable or its realization \(y^i\)). 
Since all latent and variational distributions $q$ have either fixed hyper-parameters or parameters learned from data, we drop sub-script $\theta$.
The expressivity of a normalizing flow is enhanced by composing multiple simple diffeomorphisms.
Rather than using a single transformation \(\mathscr{f}\), one can define:

\begin{align}
    \mathscr{f} & = \mathscr{f}^L \circ \mathscr{f}^{L-1} \circ \cdots \circ \mathscr{f}^1 \label{eq:flow_composition}
\end{align}

Where each \(\mathscr{f}^i: \mathcal{M}^{i-1} \to \mathcal{M}^{i}, \ i\in \{1:L\}\) is a diffeomorphism. 
This composition allows the overall transformation to be more flexible while retaining the invertibility and smoothness properties required for density evaluation. 
The change of variables formula then becomes:

\begin{align}
q(y^L) & = q(y^{0}) \cdot \prod_{i=1}^L \abs \left( \left| \pmb{J}^{\mathscr{f}_i}(y^{1-1}) \right| \right)^{-1} 
\end{align}

Since \( L \in \mathbb{N}^*\), the models described above are discrete normalizing flows.

\newpage

\subsection{Continuous normalizing flows \& flow matching on Riemannian manifolds}

In the continuous limit the discrete sequence of transformations is replaced by a continuous evolution governed by an ordinary differential equation (ODE). 
This yields continuous normalizing flows (CNFs) or neural ODEs \cite{chen_neural_2019, grathwohl_ffjord_2018}, which have been extended to Riemannian manifolds \cite{mathieu_riemannian_2020}. 
In this setting, we consider a time-dependent flow \(h: \mathcal{M} \times [0,1] \to \mathcal{M}\) and a time-dependent velocity field \(\psi: T\mathcal{M} \times [0,1] \to T\mathcal{M}\).
\(T\mathcal{M}\) is the tangent bundle of \(\mathcal{M}\), and the velocity field is modeled by a neural network. 
We denote the evolving state at time \(t\) by \(h^t \equiv h(y, t)\).
The time-evolution of the flow is defined as follows:

\begin{align}
\frac{d\,h^t}{dt} &= \psi(h^t, t; \theta), \quad h^0 \sim q(H^0) \label{eq:flow_with_initial}
\end{align}

Equation~\eqref{eq:flow_with_initial} governs the evolution of the state starting from the base distribution \(q(H^0)\). 
In the continuous setting we assume that \(\mathcal{M}^0 = \mathcal{M}^1 = \mathcal{M}\), where the manifold is embedded in ambient space \(\mathbb{R}^{K+1}\). 
We equip \(\mathcal{M}\) with a Riemannian metric:

\begin{align}
g: \quad T_h\mathcal{M} \times T_h\mathcal{M} &\to \mathbb{R}_+ \label{eq:riemannian_metric}
\end{align}

Which assigns an inner product to the tangent space at each point $h\in \mathcal{M}$. 
The evolution of the log-density along the flow is given by the continuity equation:

\begin{align}
\frac{d\,\ln q(h^t)}{dt} &= - \nabla_g \cdot \psi(h^t, t; \theta) \label{eq:continuity_equation}\\
&= - \text{Tr}\!\Bigl(g^{-1}(h^t)\frac{\partial \psi(h^t, t; \theta)}{\partial h^t}\Bigr)\label{eq:trace}
\end{align}

$\text{Tr}(...)$ in Equation~\eqref{eq:trace} denotes the trace of the Jacobian of the vector field.
Integration over time yields the probability path:

\begin{align}
q(h^t) &= q(h^0) \cdot \exp\!\Bigl(-\int_0^t \nabla_g \cdot \psi(h^s, s; \theta) \, ds\Bigr)
\end{align}

We require that the probability path satisfies two boundary conditions.
At \(t=0\) the path is equal to the base distribution $q(H^0) = q(Y^0)$ and at \(t=1\), the path approximates the target distribution $q(H^1) \approx p(Y)$.

This continuous perspective not only provides a smooth and adaptive way to model transformations but also avoids the need for expensive determinant computations allowing for transformations with non-triangular Jacobians to be used. 
The trade-off is that for CNF training an ODE must be integrated for pairs of samples from the base and target distributions to compute a KL divergence loss which requires the numerical evaluation of the Jacobian trace, which can become a computational bottleneck in high-dimensional settings.
The canonical solution to this computational bottle-neck is to use Hutchinsons trace estimator \cite{grathwohl_ffjord_2018}.

A promising alternative to training CNFs is provided by flow matching \cite{lipman_flow_2024, lipman_flow_2023}, which directly aligns \(\psi\) with a target vector field \(\psi^*\) without requiring ODE integration during training. 
Flow matching too applies to Riemannian manifolds \cite{chen_flow_2024}.
Flow matching starts by defining the marginal probability path in terms of a conditional probability path:

\begin{align}
q(h^t) &= \int_{\mathcal{M}} q(h^t|y) \, p(y) \, d\vol (y) \label{eq:marg_prob_path}
\end{align}

Where \(d\vol (y)\) denotes the Riemannian volume measure on \(\mathcal{M}\) and $y$ is just a single data point.
In conditional flow matching, one defines a target conditional probability path \(q(H^t|y)\) with the following boundary conditions.
At \(t=0\), samples follow the base distribution \(q(H^0|y) = p(y^0)\)
and at \(t=1\), all probability is concentrated at a single data point: \(q(H^1|y) = \delta_{y}(H^1)\), where $\delta_{y}$ is the Dirac delta function centered at data point $y$.
The conditional velocity field that generates path \ref{eq:marg_prob_path} is then given by:

\begin{align}
\psi^*(h^t) &= \int_{\mathcal{M}} \psi^*(h^t|y) \, \frac{q(h^t|y)\, p(y)}{q(h^t)} \, d\vol (y)
\end{align}

The conditional velocity field is the time derivative of the conditional flow:

\begin{align}
\frac{d\,h^*(t | y)}{dt} &= \psi^*(h^t|y)
\end{align}

A natural choice for the target conditional flow is the geodesic conditional flow:

\begin{align}
h^*(t) &= \exp_{h^0}\!\Bigl(t\,\ln_{h^0}(h^1)\Bigr), \quad t\in[0,1]
\end{align}

Which guarantees that \(h^*(0)=h^0\) and \(h^*(1)=h^1\). 
Here, \(\exp_{h^0}\) denotes the Riemannian exponential map at \(h^0\), and \(\ln_{h^0}(h^1)\) is its inverse, the logarithmic map. 
In Riemannian geometry, geodesics are the curves of shortest distance between points on the manifold.
Thus, choosing the geodesic path ensures that the flow transports the density optimally from the base distribution to the target distribution, respecting the intrinsic geometry of the manifold. 
The Riemannian conditional flow matching (RCFM) loss is defined by:

\begin{align}
\mathcal{L}_{\text{RCFM}}(\theta) &= \mathbb{E}_{t \sim \mathcal{U}(0,1),\, h^0 \sim q(H^0|y),\, y\sim p(Y)} \Bigl\Vert \psi(h^t, t; \theta) - \psi^*(h^t|y) \Bigr\Vert_G^2
\end{align}

where \(\psi^*(h^t|y)\) is the target vector field that guides the flow along the geodesic between \(h^0\) and \(h^1\).

At inference time, a sample is drawn from the base distribution and the trained flow is integrated until \(t=1\) to yield a sample from the approximate target distribution. 
In conditional Riemannian flow matching, however, the vector field \(\psi\) is defined on the ambient space \(\mathbb{R}^{K+1}\) rather than directly on the manifold \(\mathcal{M}\). 
As a result, the numerical integration of the ordinary differential equation (ODE) may yield points that deviate from the manifold. 
To ensure that the integration remains on the manifold, we apply a projection operator at each step:

\begin{align}
\pi(x) &= \arg \min_{y\in \mathcal{M}} \|x - y\|_g \label{eq:manifold_projection}
\end{align}

This projection operator maps any point \(x \in \mathbb{R}^{K+1}\) to the closest point \(y \in \mathcal{M}\) in terms of the Riemannian distance \(\|\cdot\|_g\). 
In practice, this retraction step is essential for maintaining the fidelity of the flow to the manifold's geometry, compensating for potential numerical errors or the drift caused by operating in the higher-dimensional ambient space.
In summary, we choose the geodesic path because it represents the optimal transport of density between \(H^0\) and \(h^1\) on the manifold. 
The projection operator is necessary to enforce the manifold constraint during ODE integration, ensuring that the evolution of the flow remains on \(\mathcal{M}\) at every step.

\subsection{Our contribution}

We will show show that any full-dimensional polytope is homeomorphic to a unit ball of the same dimension. 
We subsequently show that learning a target distribution on a ball is sufficient to model distributions on a polytope. 
We rely on the circular spline flows introduced in \textcite{rezende_normalizing_2020} to model distributions on the sphere and extend them to distributions over the ball and polytope by simply including a radius from the origin.
Since a polytope is a Riemannian manifold, we show how one can use Riemannian CNFs \cite{mathieu_riemannian_2020} combined with flow matching \cite{lipman_flow_2023, chen_flow_2024} to model distributions on the ball and the polytope. 
Finally, we propose a strategy to construct flows using only the V-representation, thereby avoiding the computational cost of converting to the H-representation.

\newpage

\section{Methods}

We begin by showing how to transform any given polytope into its corresponding full-dimensional John polytope which is centered at the origin and whose facets all touch the unit ball \( \mathbb{B}^K(1)\) \cite{john_extremum_2014}.
Subsequently, we introduce the mapping between this John polytope and the unit ball. 
In the last three subsections, we introduce how to model discrete flow on polytopes given the H-representation and CNFs on polytopes given either the H or V representation.

\subsection{Transforming and rounding a polytope}
\label{sec:trans_n_round}

Given a polytope in H-representation, it is typically not full-dimensional.
Therefore, we first need to find the minimal affine subspace of the polytope, as shown in Section 2.2.1 of \textcite{liphardt_efficient_2018}:

\begin{align}
    v^\ddagger &= \pmb{T}\, v^\dagger + \tau \quad \forall\, v^\ddagger \in \mathcal{F}^\ddagger, \quad \pmb{T}\in \mathbb{R}^{R \times K},\quad K \le R \label{eq:free_vars}
\end{align}

Equation~\eqref{eq:free_vars} expresses every point in the polytope in terms of free variables of dimension $K$. 
To determine the embedding parameters \(\pmb{T}\) and \(\tau\), the original polytope is simplified by (1) removing redundant constraints, via solving two LPs per inequality, and (2) collecting implicit equality constraints from \(A_c\) into \(S^+\).
This matrix can be thought of as an extended equality constraint matrix with extra rows representing constraints that were previously 'hidden' in matrix \(\pmb{A}_c\).
There are then two choices for the kernel \(\ker(\pmb{S}^{+})\):

\begin{align}
    v^{0} & = \argmin_{v^{0}} \max_{v^\ddagger \in \mathcal{F}^\ddagger} \| v^\ddagger - v^{0}\|_2^2 \label{eq:chebyshev_center}\\[1mm]
    & 
    \begin{cases}
    \pmb{T}^{rref} = 
    \begin{bmatrix}
        \pmb{I} \\[1mm]
        \pmb{T}^\star
    \end{bmatrix} = \frac{\partial v^\ddagger}{\partial v^\dagger} \quad \text{s.t.} \quad v^\ddagger = \begin{bmatrix}
        v^\dagger \\[1mm]
        v^{\star}
    \end{bmatrix}
    \\[1mm]
    \tau^{rref} = v^{0} - \pmb{T}^{rref} \, v_{:K}^{ 0}
    \end{cases}
    \label{eq:rref_kernel}\\[1mm]
    & 
    \begin{cases}
    S^+ = \pmb{U}\, \pmb{\Sigma}\, \pmb{V}^T,\quad \pmb{T}^{SVD} = \pmb{V}_{:,-K:}
    \\[1mm]
    \tau^{SVD} = v^{ 0} 
    \end{cases} 
    \label{eq:svd_kernel}
\end{align}

Equation~\eqref{eq:chebyshev_center} defines the Chebyshev center of the polytope, while Equations~\eqref{eq:rref_kernel} and \eqref{eq:svd_kernel} describe two embedding strategies: the row reduced echelon form (RREF) embedding and the singular value decomposition (SVD) embedding, respectively.
With a subscript, we denote the first $K$ elements of a vector as $v_{:K}$ and the last $K$ columns 
of matrix as $\pmb{V}_{:,-K:}$.
We denote dependent variables with the superscript \(v^\star\), which indicates that the free variables in the $rref$ embedding are a subset of the original variables \(v^\ddagger\).
The transformed polytope, expressed in the free-variable space, is then given by

\begin{align}
    \mathcal{F}^{\dagger} &= \{ v^\dagger \in \mathbb{R}^K \mid \pmb{A}^\dagger\, v^\dagger \le b^\dagger \} \label{eq:full_dim_pol} \\
    & \quad \text{with } \pmb{A}^\dagger  = \pmb{A}^\ddagger\, \pmb{T} \text{ and } b^\dagger = b^\ddagger - \pmb{A}^\ddagger\, \tau \nonumber
\end{align}

Equation~\eqref{eq:full_dim_pol} represents the full-dimensional polytope in the free-variable space.
In the \ce{$^{13}$C}-MFA literature the RREF embedding is common since the free variables share the same units as the original fluxes \cite{quek_openflux_2009}, whereas the SVD embedding is used more generally when working with polytopes \cite{haraldsdottir_chrr_2017, zhang_numerical_2003}.
Rounding the transformed polytope \(\mathcal{F}^\dagger\) involves transforming it to maximum isotropic (or John) position by minimizing the sandwiching ratio $\frac{\hat{\Phi}}{\hat{\phi}}$, where:

\begin{align}
    \hat{\Phi} & = \argmax_{\Phi} \mathbb{B}^K(\Phi) \supseteq \mathcal{F}\label{eq:contained_ball}\\
    \hat{\phi} & = \argmin_{\phi} \mathbb{B}^K(\phi) \subseteq \mathcal{F}\label{eq:containing_ball}
\end{align}

Equations~\eqref{eq:contained_ball} defines the largest ball contained within the polytope and Equation~\eqref{eq:containing_ball} is the smallest ball containing the polytope.
To minimize the sandwiching ratio, let us define an ellipsoid that is centered at \(\epsilon\) as follows:

\begin{align}
    \mathcal{E} &= \{ v^\dagger \in \mathbb{R}^K \mid (v^\dagger - \epsilon)^T\, (\pmb{E}\, \pmb{E}^T)^{-1}\, (v^\dagger - \epsilon) \le 1 \} \nonumber\\ 
    & \quad \text{with } \pmb{E} = \pmb{E}^T \text{ and } y\, \pmb{E}\, y^T \ge 0 \ \forall\, y \in \mathbb{R}^K\\[1mm]
    &= \{ v^\dagger \in \mathbb{R}^K \mid v^\dagger = \pmb{E}\, y + \epsilon,\quad y \in \mathbb{B}^K(1) \} \label{eq:ellipsoid}
\end{align}

Equation~\eqref{eq:ellipsoid} defines a $K$-dimensional ellipsoid centered at \(\epsilon\) with the matrix \(\pmb{E}\) satisfying the required positive semi-definite condition.
Finding the maximum volume ellipsoid (MVE) contained in \(\mathcal{F}^\dagger\) is a convex optimization problem that maximizes the determinant of \(\pmb{E}\) and can be solved efficiently as shown in \cite{zhang_numerical_2003}.
The free variables $v^\dagger$ can then be expressed in terms of rounded variables $v$ as follows:

\begin{align}
    v^\dagger & = \pmb{E}v + \epsilon, \quad \forall \, v^\dagger \in \mathcal{F}^\dagger \label{eq:rounded_vars}
\end{align}

The affine transformation that maps points from the \(K\)-dimensional unit ball to this maximum volume ellipsoid is then used to construct the John polytope:

\begin{align}
    \mathcal{F} &= \{ v \in \mathbb{R}^K \mid \pmb{A}\, v \le b  \} \label{eq:John_pol}
    \\[1mm]
    & \quad \text{with } \mathbb{B}(1)^K \subseteq \mathcal{F} \subseteq \mathbb{B}(\Phi)^K \text{ for some } \Phi \ge 1 \nonumber
    \\[1mm]
    & \quad \text{and } \pmb{A} = \pmb{A}^\dagger\, \pmb{E},\quad b = b^\dagger - \pmb{A}^\dagger\, \epsilon \nonumber
\end{align}

We model distributions over the polytope using the rounded variables \(v\). 
Because \(\pmb{T}\) in Equation~\eqref{eq:free_vars} has full column rank and \(\pmb{E}\) in Equation~\eqref{eq:rounded_vars} is square and invertible, both mappings are bijective. 
This bijectivity guarantees that we can uniquely map back to the original polytope variables \(v^\ddagger\).

\subsection{The hit-and-run ball transform}
\label{sec:hit_run_trans}

\begin{wrapfigure}[10]{r}{0.4\textwidth}
    \centering
    \includegraphics[width=0.336\textwidth]{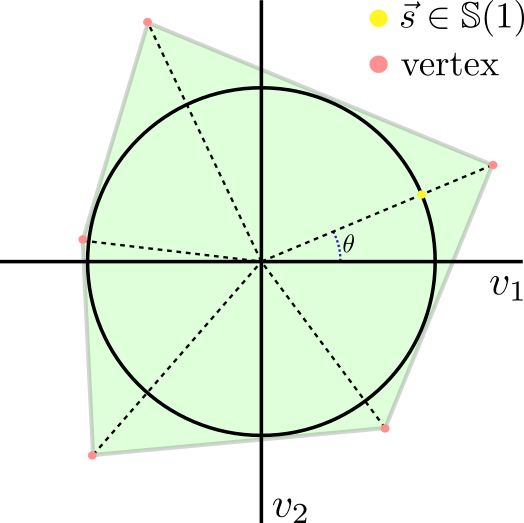}
    \caption{\textit{Graphical intuition for the homeomorphism between a 2D convex polytope and the disk.}}
    \label{fig:projection_intuition}
\end{wrapfigure}

The transformation from the John polytope to the ball is the inverse of transformation that is used in the hit-and-run (HR) algorithm for the sampling of polytopes \cite{smith_efficient_1984, kaufman_direction_1998}.
A modified HR algorithm is described in Appendix \ref{appendix:multi_prop_hr_sampling} which is used in the experiments of Section \ref{sec:experiments} to sample from target densities over a polytope. 
For a point $v\in \mathcal{F}$, the ball transformation consists of the following steps:

\begin{align}
    d & = \| v \|_2 \label{eq:eucl_norm}
    \\
    s & = \frac{v}{d} \ \text{with: } s \in \mathbb{S}^{K-1}(1)
    \\
    \alpha & = b \oslash (\pmb{A}\, s) \label{eq:A_dist}
    \\
    \alpha^{max} & = \min(\alpha \mid \alpha \ge 0) \label{eq:compute_alpha}
    \\
    r & = \Bigl( \frac{d}{\alpha^{max}} \Bigr)^{\frac{1}{K}},\quad r \in [0,1] \label{eq:k_norm}
    \\
    \beta & = r \cdot s,\quad \beta \in \mathbb{B}^K(1) \label{eq:ball_coords}
\end{align}

Where $d$ is the Euclidean norm of point $v$ and $s$ denotes the cartesian coordinates for a point on the unit sphere, which can be thought of as the direction from the origin along which point $v$ lies.
The distances to all half-planes are computed via Equation~\eqref{eq:A_dist} and the distance to the closest constraint is $\alpha^{\max}$.
Next, a scaled \(K\)-norm is defined in Equation~\eqref{eq:k_norm} and the Cartesian ball coordinates are obtained in Equation~\eqref{eq:ball_coords}.
Equations \eqref{eq:eucl_norm} through \eqref{eq:ball_coords} define the homeomorphism:

\begin{align}
    \mathscr{b} &: \mathcal{F} \to \mathbb{B}, & v \mapsto &\beta \label{eq:b_homeo}
    \\
    \mathscr{b}^{-1}& :  \mathbb{B} \to \mathcal{F}, & \beta \mapsto & v = \alpha^{max}\cdot \beta
\end{align}

Note that in Equation~\eqref{eq:k_norm} the exponent \(\frac{1}{K}\) can be interpreted as a tunable parameter.
An exponent closer to 1 concentrates more of the volume of the polytope towards the origin of the ball whereas an exponent closer to 0 will concentrate more volume towards the boundary of the ball.

Though \(\mathscr{b}\) is a homeomorphism, it is not a diffeomorphism since the function is not differentiable everywhere.
Along directions where a chord points towards a \(K-n\)-face where \(K>n>1\), i.e. a vertex (0-face) or an edge (1-face), the \(\max\) function of Equation \eqref{eq:k_norm} is not defined, since multiple hyper-planes are at exactly equal distance.
This yields the discontinuity in the derivative.
Luckily, the set of points for which the derivative is not defined has measure 0, so this will likely not matter in practice.

Figure \ref{fig:projection_intuition} shows the John polytope and the inscribed unit sphere for a 2D polytope.
Transformation \(\mathscr{b}\) can be thought to squeeze all points along a chord starting at the origin to be inside the unit disk.
The yellow point lies on a chord that points towards a vertex of the polytope.
The derivative of transformation \(\mathscr{b}\) is not defined along this chord as indicated by the dotted lines.
The angle \(\theta\) in the figure is the polar coordinate of the yellow point, which in Cartesian coordinates is described by 2 numbers \(v_1, v_2\).

\subsection{Spline flows on the ball}
\label{sec:spline_flow}

We use the recursive construction from Section 2.3.1 of \cite{rezende_normalizing_2020} to model flows on the ball. 
For circular spline flows on spheres, a point \(s\) on a sphere is mapped to a point on a cylinder via the recursive application of the following transformation:

\begin{align}
    \cyl^D : \mathbb{S}^D  \times  [-1,1]^{K-D} &\to \mathbb{S}^{D-1} \times [-1, 1]^{K-D+1} \nonumber
    \\[1mm]
    c^{D} & \mapsto  c^{D-1} = \Bigl[ \frac{c_{1:D-1}}{\sqrt{1 - c_{D}^2}},\ c_{D:K}\Bigr]^T \label{eq:cylinder}
\end{align}

By composing these mappings, the overall homeomorphism is:

\begin{align}
    \cyl: \mathbb{S}^K & \to \mathbb{S}^1 \times [-1, 1]^{K-1},   & s \mapsto & c = c^{ 3:K} = \cyl^3 \circ \cdots \circ \cyl^K (s) \label{eq:cyl_map}
\end{align}

After applying function \(\cyl^3\), the first two coordinates lie on \(\mathbb{S}^1\) and we transform these into a single spherical coordinate \(\theta = \atan2 (c_1, c_2)\). 
In figure \ref{fig:projection_intuition}, \(\theta\) is indicated as the angle w.r.t. the positive x-axis.
The resulting cylinder vector with polar coordinate is thus: \(\varphi = [\theta, c_{3:K}, r]\), where $r$ is the radius from Equation~\eqref{eq:k_norm}.
We can model distributions over \(\theta\) using circular spline flows (section 2.1.2 of \cite{rezende_normalizing_2020}) and use regular spline flows \cite{durkan_cubic-spline_2019, durkan_neural_2019} to model distributions over \(r\) and \(c_{3:K}\).
The resulting density of a point in the polytope is given by:

\begin{align}
    q^{spline}(v) & = p_\square^{\mathcal{U}}(\varphi) \cdot | \pmb{J}^{spline} |^{-1} \cdot | \underbrace{ \pmb{J}^{v \beta} \, \pmb{J}^{\beta c} \, \pmb{J}^{c \varphi} }_{\pmb{J}^{v \vartheta}} |^{-1}
\end{align}

The base distribution, \(p_\square^{\mathcal{U}}(\varphi)\), is defined as a uniform density (denoted by \(\mathcal{U}\)) over a hyper-rectangle \(\square\). 
Its bounds are given by \([-\pi, \pi] \times [-1,1]^{K-2} \times [0, 1]\), which correspond to the ranges for \(\theta\), \(c_{3:K}\), and \(r\), respectively.
We then define several Jacobians for the various coordinate transformations. 
First, \(\pmb{J}^{spline}\) is the Jacobian of the spline-based transformations, and it is triangular due to its autoregressive structure. 
\(\pmb{J}^{v \beta}\) is the Jacobian corresponding to the mapping from ball coordinates to polytope coordinates as specified in Equation~\eqref{eq:b_homeo}. 

Similarly, the Jacobian \(\pmb{J}^{\beta c}\) is a \(K \times (K+1)\) matrix for the transformation from cylinder coordinates to ball coordinates, and \(\pmb{J}^{c \varphi}\) is the \((K+1) \times K\) Jacobian that converts from polar to Cartesian cylinder coordinates. 
The mapping from polar to Cartesian coordinates is a diffeomorphism.
The cylinder mapping, Equation~\eqref{eq:cyl_map}, is a homeomorphism with a measure 0 set of points where the mapping is not differentiable \cite{rezende_normalizing_2020}, similar to $\mathscr{b}$ of Equation~\eqref{eq:b_homeo}.

The combined transformation from polar cylinder to polytope coordinates is represented by the square Jacobian \(\pmb{J}^{v \vartheta}\), which is dense and contains no learnable parameters.
In our implementation, when density estimation is required, we compute the determinant of \(\pmb{J}^{v \vartheta}\) directly for all samples in a batched fashion without relying on automatic differentiation. 
This strategy allows for efficient parallelization and improves the performance of this computationally intensive step.

\subsection{Riemannian continuous flows and flow matching on polytopes and balls}
\label{sec:flow_matching}

While our initial work focused on discrete normalizing flows, continuous flows trained with flow matching have become the dominant paradigm in flow-based modeling. 
One challenge with CNFs is that, during inference, integrating the underlying ODE may yield solutions that fall outside the target polytope. 
To counteract this, variables are projected back onto the manifold at each integration step using the projection operator defined in \eqref{eq:manifold_projection} \cite{chen_flow_2024}. 
For polytopes, this projection requires solving a quadratic program, the formulation of which depends on whether the polytope is represented in its H-representation or V-representation:

\begin{align}
    \pi(y) & = \arg \min_{v \in \mathcal{F}} \|y - v\|_2^2 \label{eq:quad_proj_prog}
    \\
    & = \arg \min_{\pmb{A}\cdot v \leq b} \|y - v\|_2^2  \notag
    \\
    & = \arg \min_{\substack{ \lambda \in \mathbb{\Delta}_1}} \|y - \pmb{V}\, \lambda\|_2^2 \notag
\end{align}

Solving a quadratic program at every ODE iteration, whenever the solution exits the polytope, is computationally demanding. 
In our PyTorch implementation, we encountered a limitation: PyTorch currently does not support parallelized gradient tracking through functions that involve data-dependent control flow (see \href{https://github.com/pytorch/functorch/issues/257}{github issue}). 
As a result, we cannot monitor the convergence of the quadratic program in parallel and must resort to sequentially evaluating each sample in a batch.
This approach is considerably slow, even for low-dimensional models.

A Euclidean flow is defined in unconstrained Euclidean space using a standard Euclidean metric, so it does not inherently enforce the boundary constraints of a polytope. 
However, when both the base and target distributions are confined to a polytope, a perfectly matched Euclidean flow, i.e. one for which \(KL(q^{eucl} \| p(y)) = 0\), should, in principle, never generate samples outside the polytope. 
Consequently, we investigate Euclidean flows that do not incorporate any projection back onto the manifold. 
The density of such a flow is evaluated as follows:

\begin{align}
    q^{eucl}(v) & = \begin{cases}
        p_\mathcal{F}^{\mathcal{U}}(v) \cdot \exp \Bigl( - \int_0^1 \nabla \psi(s) ds \Bigr), & v \in \mathcal{F}\\[1mm]
        0, & v \notin \mathcal{F}
    \end{cases}
\end{align}

Where $p_\mathcal{F}^{\mathcal{U}}$ is the uniform base density over the polytope.
The $q^{eucl}(v) = 0 \ \forall \ v \notin \mathcal{F}$ is necessary to reject samples that left the polytope.
Note that for Euclidean flows the integrand of the divergence term is simply: \(\nabla \psi(s)\), since no extra computation is required for Euclidean metric \(g\).

Assuming that we have access to the H-representation of the polytope, we can again model continuous flows on the polytope using flows on the ball. 
We choose to equip the unit ball with a Euclidean metric.
Alternatively, one can use the Poincare metric, but for most applications, there is no obvious reason to do so and we did not investigate this option further.
To project points onto this ball, all that is required is a simple scaling operation: 

\begin{align}
    \pi(y) & = \frac{y}{\max (1, \|y\|_2)}
\end{align}

Though it is possible to include the boundary of the ball in Riemannian flow matching, Section G.2 of \textcite{chen_flow_2024}, we restrict our attention to the open ball, which corresponds to the open polytope. 
The density of a strictly interior point in the polytope can be evaluated as follows:

\begin{align}
    q^{ball}(v) & = p_\mathbb{B}^{\mathcal{U}}(\beta) \cdot \exp \Bigl( - \int_0^1 \nabla_g \psi(s) ds \Bigr) \cdot \Bigl| \pmb{J}^{v\beta} \Bigr|^{-1}
\end{align}

Where ball coordinate \(\beta \in \mathbb{B}\) and a base distribution defined as a uniform density over the unit ball, denoted \(p_\mathbb{B}^{\mathcal{U}}\).
Jacobian $ \pmb{J}^{v\beta}$ is square and thus not equal to Jacobian $\pmb{J}^{v\beta}$ in section \ref{sec:spline_flow}, where radius $r$ was modeled as a separate variable.

For exact density evaluation, we need to be able to evaluate the normalized density of the base distribution. 
For the ball flow, it is trivial to choose a distribution whose probability can be evaluated analytically. 
Conversely, for the Euclidean flow, the density can typically only be approximated through sampling.
For example, the uniform density for a sample in the polytope is: \(p_\mathcal{F}^{\mathcal{U}}(v) = \frac{1}{\vol \mathcal{F}}\) and for higher dimensions, the volume of the polytope must be approximated through sampling.

\newpage

\subsection{Flows on barycentric coordinates}

In Sections \ref{sec:trans_n_round} -- \ref{sec:flow_matching}, we assumed to have access to the H-representation of a polytope.
If instead, we imagine to only have access to the V-representation of a polytope, we would like to still be able to model distributions over it.
Converting from the V to the H representation of a polytope has a time-complexity of \(O((V+1)^{\floor{\frac{K}{2}}})\) \cite{avis_pivoting_1992}.
This conversion is therefore typically not computationally feasible for polytopes with many vertices or in higher dimensions unless a trivial description is available, e.g. for regular polytopes like hyper-rectangles, simplices or cross-polytopes. 
For this Section, we consider a full-dimensional polytope in V-representation and assume we have samples from a distribution over this polytope. 

The barycentric coordinates for a point on a simplex are unique since \(V=K\). 
For general polytopes, it is typically the case that \(V \gg K\) which makes the system \(\pmb{V}\cdot \lambda = v\) under determined, meaning that the mapping $\mathcal{F} \rightrightarrows \Delta_1^V$ is a set-valued function and thus not bijective.
We can therefore not apply the probabilistic change of variables of Equation \eqref{eq:change_of_vars}.
This can be solved by choosing a unique barycentric coordinate mapping.
A computationally feasible choice are the maximum entropy coordinates ($\mec$) \cite{hormann_maximum_2008}:

\begin{align}
    \mec&: \mathcal{F} \to \Lambda, & v  \mapsto & \lambda^{\mec} = \arg \max_{\substack{\lambda \in \Delta_1 \\ \pmb{V}\, \lambda = v}} -\lambda^T \ln (\lambda) \label{eq:max_ent_coor}
    \\[1mm]
    \mec^{-1}& : \Lambda \to \mathcal{F}, & \lambda^{\mec} \mapsto & v = \pmb{V}\, \lambda^{\mec}\label{eq:inv_max_ent_coor}
\end{align}

With 

\begin{align}
    \Lambda & = \{\lambda^{\mec} = \mec (v) \mid v \in \mathcal{F}\}
\end{align}

For brevity, we will drop the $\mec$ superscript from \(\lambda\) in what follows.
Note that the $\mec$ is defined only for points in the open polytope, because at the boundary some coordinates on the simplex become zero, and \(\ln(0)\) is undefined.

Equations~\eqref{eq:max_ent_coor} and \eqref{eq:inv_max_ent_coor} define the $\mec$ mapping and its inverse, which is bijective by construction. 
As demonstrated in \textcite{hormann_maximum_2008}, the forward $\mec$ mapping is smooth over the open polytope, while Equation~\eqref{eq:inv_max_ent_coor} is linear and thus smooth.
Since the open \(K\)-dimensional polytope is a Riemannian manifold and the $\mec$ mapping along with its inverse are smooth everywhere, we conclude that the set \(\Lambda\) forms a smooth connected \(K\)-dimensional sub-manifold embedded in the simplex \(\Delta_1^V\) \cite{lee_submersions_2012}.

A metric on the simplex is provided by the Aitchison geometry \cite{aitchison_statistical_1982, greenacre_aitchisons_2023}.
In this framework, the isometric log‐ratio ($\ilr$) transform \cite{egozcue_isometric_2003} is defined as follows:

\begin{align}
    \ilr &: \Delta_1^V \to \mathbb{R}^{V}, & \lambda \mapsto & z = \pmb{H}\, \ln (\lambda) \label{eq:ilr_map}
    \\
    \ilr^{-1} &: \mathbb{R}^{V} \to \Delta_1^V, & z \mapsto & \lambda = \begin{cases}
    z^{a} & = \exp (\pmb{H}^T\, z)
    \\[1mm]
    \lambda & = \frac{z^{a}}{\mathbf{1}^T\, z^{a}}
    \end{cases}\label{eq:inv_ilr_map}
\end{align}

Here, \(\pmb{H} \in \mathbb{R}^{K \times V}\) is taken to be the Helmert matrix, which provides an orthonormal basis for the subspace of \(\mathbb{R}^V\).
Both the $\ilr$ and its inverse are smooth, thus making this mapping a diffeomorphism.

Because the $\ilr$ transform is an isometry between the Aitchison geometry on the simplex and Euclidean space, the linear interpolation between any two points in \(\mathbb{R}^{K}\) corresponds to the geodesic, with respect to the Aitchison metric, between the corresponding points in \(\Delta_1^V\).
An alternative geometric structure on the simplex is provided by the Fisher–Rao metric \cite{davis_fisher_2024}, but we did not investigate this option further. 
In contrast to the Aitchison metric induced by the $\ilr$ transform, the Fisher–Rao metric is defined on the entire simplex, including its boundary.

Let

\begin{align}
\mathcal{Z} = \{z = \ilr (\lambda) \mid \lambda \in \Lambda\}
\end{align}

be the $\ilr$-transformed image of \(\Lambda\). 
Since the $\ilr$ transform is an isometry, \(\mathcal{Z}\) is a \(K\)-dimensional affine sub-space of \(\mathbb{R}^{V}\). 
One may compute an orthogonal projection using singular value decomposition (SVD). 
Let the columns of matrix \(\pmb{Z}\) represent at least $K$ $\ilr$-transformed points from $\Lambda$, one may then write:

\begin{align}
\pmb{Z} = \pmb{U}\,\pmb{\Sigma}\,\pmb{W}^T \label{eq:ilr_svd}
\\
\pmb{P} = \pmb{W}_{:,:K}^T
\end{align}

Where $\pmb{P}$ equals the first $K$ rows of $\pmb{W}^T$.
To then obtain the projected $\ilr$ points, we define the projection:

\begin{align}
    \proj: & \mathbb{R}^V \to \mathbb{R}^K, & z \mapsto & z^p = \pmb{P}\,z
    \\
    \proj^{-1}: & \mathbb{R}^K \to \mathbb{R}^V, & z^p \mapsto & z = \pmb{P}^T\,z^p
\end{align}

The projected $\ilr$ transformation recovers the effective \(K\)-dimensional coordinates of the $\ilr$-transformed points. 
In practice, we first transform all samples from the target distribution into $\ilr$ coordinates, and then compute a singular value decomposition (SVD) on a large batch of data to determine an optimal \(K\)-dimensional projection. 
We denote the set of projected coordinates as:

\begin{align}
    \mathcal{Z}^p = \{z^p = \proj(z) \mid z \in \mathbb{R}^{V}\}
\end{align}

Once the target coordinates have been mapped to these projected $\ilr$ coordinates, we standardize the data by subtracting the mean and dividing by the standard deviation:

\begin{align}
    \stdz: & \mathbb{R}^K \to \mathbb{R}^K, & z^p \mapsto & z^t = (z^p - \mu) \oslash \sigma
    \\
    \stdz^{-1}: & \mathbb{R}^K \to \mathbb{R}^K, & z^t \mapsto & z^p = z^s\odot \sigma + \mu
\end{align}

In this way, every sample from the target distribution is represented in projected, standardized $\ilr$ coordinates $z^t$. 
We can then model their distribution using either discrete flows or CNFs. 
In our case, we use a Euclidean CNF whose density is:

\begin{align}
    q^{ait}(v) = p_{\mathbb{R}}^\mathcal{N}(z^t) \cdot \exp\!\Bigl(-\int_0^1 \nabla \psi(s)\,ds\Bigr) \cdot | \underbrace{  \diag (\sigma) \, \pmb{P}^T \, \pmb{J}^{ilr}\, \pmb{V} }_{\pmb{J}^{vt}} |^{-1}
\end{align}

In this expression, \(p_{\mathbb{R}}^\mathcal{N}(z^t)\) denotes the Gaussian base density. 
Both the projection, \(\proj\), and the standardization, \(\stdz\), are bijective and thus diffeomorphisms. 
If we track the dimensions through each change of variables, we see that the overall Jacobian, \(\pmb{J}^{vt}\), is a square matrix. 
In particular, we have

\begin{align}
    \diag(\sigma) &\in \mathbb{R}^{K \times K} \\
    \pmb{P}^T &\in \mathbb{R}^{K \times V} \\
    \pmb{J}^{ilr} &\in \mathbb{R}^{V \times (V+1)} \\
    \pmb{V} &\in \mathbb{R}^{(V+1) \times K}
\end{align}

The product of these matrices yields \(\pmb{J}^{vt} \in \mathbb{R}^{K \times K}\). 
Its determinant precisely captures the change in density resulting from the entire sequence of transformations.

\newpage 

\section{Experiments}
\label{sec:experiments}

The \href{https://github.com/Patrickens/sbmfi_public/tree/master/arxiv_polytope}{code} for all experiments in this Section is part of the \verb|sbmfi| package in the \verb|arxiv_polytope| folder. 
We start by defining two target densities.
The support of the first density is a polytope, \(\mathcal{F}\), that takes inspiration from a model of a small metabolism such as one might encounter in \ce{$^{13}$C}-MFA. 
The equality constraint matrix of this model is:

\begin{align} 
    \pmb{S} & = \left[
        \begin{array}{c|rrrrrrrrrrrrr}
                   & c\_out & v1 & v2 & v3 & v4 & v5 & v6 & v7 & d\_out & f\_out & biomass & h\_out & a\_in \\
        \hline
A   & 0 & -1 &  0  & 0 & 0 & 0  & 0  & 0 & 0  & 0  &  0 &    0 & 1 \\
B   & 0 &  1 & -1 & -1 & 0 & 0  & 0  & 0 & 0  & 0  & -0.6 &  0 & 0 \\
C   & 0 &  0 &  0 &  1 & 0 & -1 & 0  & 0 & 0  & 0  & -0.1 &  0 & 0 \\
D   & 0 &  0 &  0 &  1 & 1 & 0  & -1 & 0 & -1 & 0  &  0 &    0 & 0 \\
E   & 0 &  1 & -1 & -1 & 1 & -1 & -1 & 0 & 0  & 0  & -0.5 &  0 & 0 \\
F   & 0 &  0 &  0 &  1 & 1 & 0  &  1 & -2 & 0  & -1 &  0 &    0 & 0 \\
H   & 0 &  0 &  0 &  0 & 0 & 0  &  0 & 1 & 0  & 0  & -0.3 & -1 & 0 \\
cof & -1 &  0 &  0 &  1 & 0 & 0  &  0 & 0 & 0  & 0  &  0 &    0 & 0 \\
        \end{array}
    \right]
\end{align}

The rows represent chemical species and the columns represent biochemical reaction rates (fluxes).
In \ce{$^{13}$C}-MFA one typically assumes \(h = 0\). 
The inequality constraints are set as \(0.05\leq v_{biomass} \leq 1.5\), \(10 \leq v_{a\_in}\leq 10\) (an implicit equality), and for the remaining variables \(0 \leq v_i \leq 100\). 
The RREF embedding (Equation~\eqref{eq:rref_kernel}) is used to determine the minimal affine subspace (Equation~\eqref{eq:free_vars}) whose dimension is \(K=4\). 
The RREF embedding yields \(v7\), \(h\_out\), \(biomass\), and \(f\_out\) as free variables, with the others affinely dependent on these. 
The transformation parameters (\(\pmb{T}\), \(\tau\), \(\pmb{E}\), \(\epsilon\)) are computed using the \verb|PolyRound| package \cite{theorell_polyround_2022}.
These parameters are shown in the Jupyter notebooks accompanying this paper.
The independent variables of the John polytope are 'rounded' and therefore obtain an \(R\_\) prefix.

We define a mixture of three Gaussians constrained to the polytope $\mathcal{F}$ described above to be our target distribution.
We know that the unit ball inscribed in the polytope touches every facet, and thus if we choose the mean of each Gaussian such that \(\| \mu^i \|_2 \approx 1\), we expect a significant fraction of the unconstrained density to land outside of the polytope.
For visualization reasons, we chose a value slightly larger than 1 and define the means of the three Gaussians:

\begin{align}
\left[
    \begin{array}{c|cccc}
                       & R\_v7         & R\_f\_out    & R\_biomass    & R\_h\_out   \\
        \hline
        \mu^1=   & -1.015        & 0            & 0             & 0           \\
        \mu^2=   & 0             & 0            & 1.015         & 0           \\
        \mu^3=   & 0             & 0            & 0             & 1.015       \\
    \end{array}
\right]
\end{align}

We denote this target distribution as \(p_\mathcal{F}^{mog}\).
For the second target density, the support is a \(K=20\) dimensional hyper-cube, denoted with overloaded notation $\square$.
Again, we use a mixture of 3 Gaussians whose means are \(\mu_i^i=1.015 \quad \forall i \in \{1:3\}\), meaning the only non-zero entry in the mean-vector is in one of the first three dimensions of the hypercube.
The weights are the same as for the polytope and the covariances are computed in the exact same way. 
We denote this target distribution as \(p_\square^{mog}\).

The density of a point in the polytope can be evaluated as follows:

\begin{align}
    p^{mog}(v) & = \sum_{i=1}^{3}\Theta_i\, \mathcal{N}(v;\mu^i, \Sigma^i) \\
    p_\mathcal{F}^{mog} (v) & = \frac{p^{mog}(v)}{\int_\mathcal{F} p^{mog}(v) \, dv} = \frac{p^{mog}(v)}{Z_\mathcal{F}}\\
    Z_\mathcal{F} & \approx \frac{\vol \mathcal{F}}{N}\sum_{i=1}^N p^{mog}(v_i),\quad v_i \sim p_\mathcal{F}^{\mathcal{U}},\quad N = 125000\\
    & \approx 0.6336 \label{eq:F_Z}
    \\
    Z_\square & \approx 0.1637 \label{eq:square_Z}
\end{align}

To sample from the mixture of Gaussian distributions with polytope support, we employ a custom multi-proposal random direction hit-and-run sampler (see Appendix~\ref{appendix:multi_prop_hr_sampling}), which also details the implementation and convergence statistics for both MCMC algorithms used for these densities.

\begin{figure}[h]
    \centering
    \includegraphics[width=0.7\textwidth]{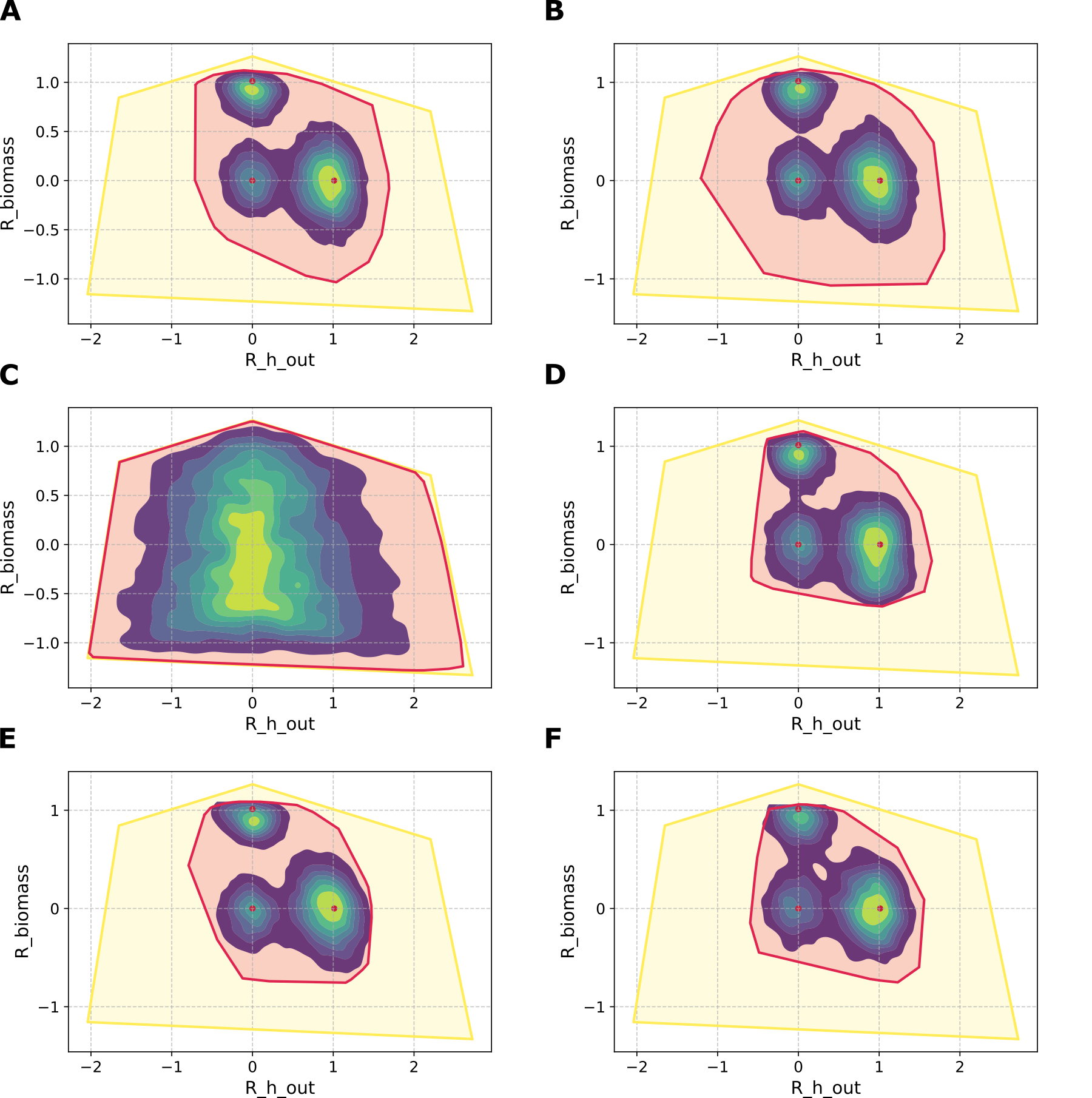}
    \caption{Yellow: the John polytope (projected onto 2D). Red: convex hull of all samples from either the flow or target. Kernel density estimates (from 10k samples) are overlaid, with red points indicating the means of the three Gaussians in the target density \(p^{mog}_\mathcal{F}\). (A) 105k samples from MCMC; (B) 20k samples from a cylinder spline flow \(q^{spline}\); (C) 125k samples from the uniform target \(p^{\mathcal{U}}_\mathcal{F}\) (via MCMC); (D) 20k samples from a Euclidean CNF \(q^{eucl}\); (E) 20k samples from a Riemannian CNF \(q^{ball}\); (F) 20k samples from a Euclidean CNF on standardized, projected $\ilr$ coordinates \(q^{ait}\).}
    \label{fig:flows_fig}
\end{figure}

For the Euclidean CNFs, we use samples from the uniform density over both the polytope \(\mathcal{F}\) and the hyper-cube as the base density. 
The polytope \(\mathcal{F}\) has dimensionality \(K=4\), which is sufficiently low to enable the exact computation of its volume using the Qhull algorithm \cite{barber_quickhull_1996}. 
In contrast, the hypervolume of the hyper-cube is available analytically.

For both target densities, we estimate the normalizing constants \(Z_\mathcal{F}\) and \(Z_\square\) using samples from the corresponding uniform distributions, with the estimated values provided in Equations~\eqref{eq:F_Z} and \eqref{eq:square_Z}, respectively. 
Notably, for \(\mathcal{F}\), a little over one-third of the unconstrained mixture density lies outside of the polytope.

\begin{figure}[h]
    \centering
    \includegraphics[width=0.7\textwidth]{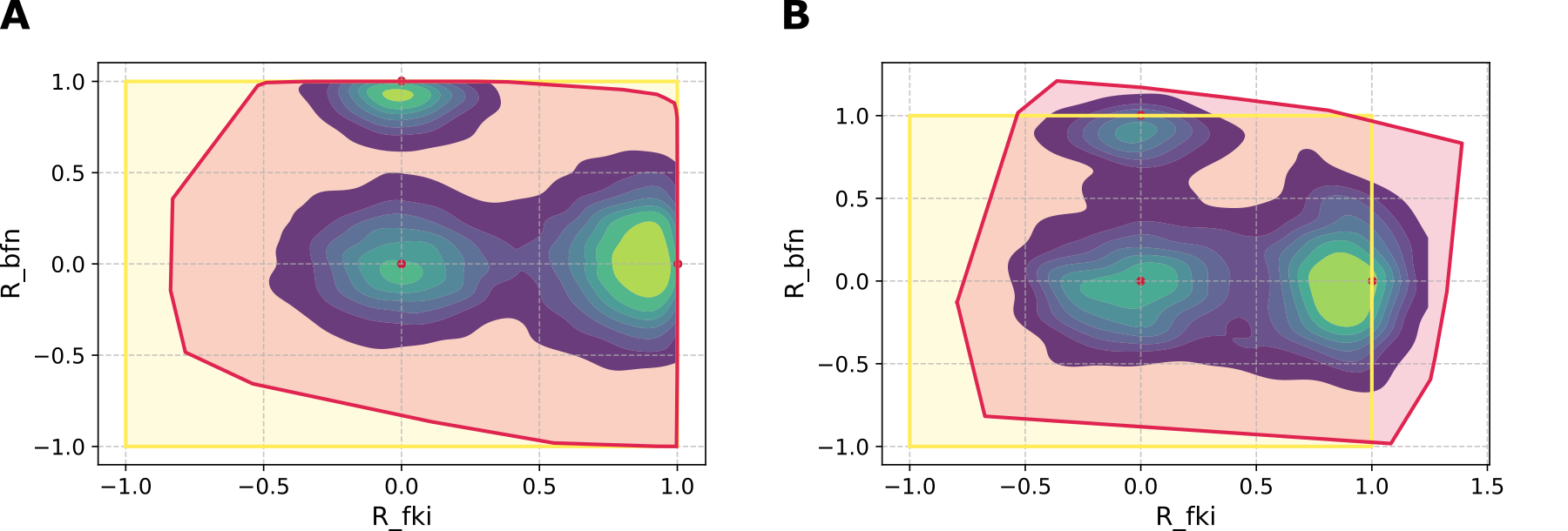}
    \caption{Similar to Figure~\ref{fig:flows_fig}, but here \(p^{mog}_\square\) is the target. (A) 125k MCMC samples; (C) 20k samples from the Euclidean CNF \(q^{eucl}\).}
    \label{fig:Kflows}
\end{figure}

\newpage

For \(q^{spline}\), \(q^{ball}\), and \(q^{ait}\), the overall transformation involves a sequence of variable mappings (e.g. via \(\pmb{J}^{\beta c}\) or \(\pmb{J}^{ilr}\)), whose composite Jacobian determinant must be computed. 
Since at least one Jacobian in these sequences is dense, the cost of computing the determinant scales as \(O(K^3)\). 
To mitigate this computational burden, we train \(q^{spline}\) using the forward KL divergence with samples drawn from the target density after transforming them onto the manifold of the base density. 
For CNFs, the support of the target distribution must match that of the base; however, this is not the case for \(q^{ball}\) and \(q^{ait}\), where a CNF is combined with a fixed (non-learnable) transformation that maps the distribution to another manifold, similar to the approach in discrete flows. 
Moreover, since evaluating the Jacobian for all samples via automatic differentiation is resource intensive, we compute all Jacobians for fixed transformations manually in a batched fashion.

In order to compare different flows, we use the effective sample size (ESS) as a percentage fraction of the number of samples from the trained flow and the Kullback-Leibler (KL) divergence of the trained flow. 
For the Euclidean flows, we also report the fraction of samples from the trained flow that fall outside of the polytope. 
Let \(\mathcal{S}\) denote a set of \(S\) samples from a flow.
The flow evaluation metrics are then computed as follows:

\begin{align}
    w_i & = \dfrac{p(x_i)}{q(x_i)} \quad \forall i \in \{1:S\}
    \\
    KL(q\|p) & = \mathbb{E}_{q}[\ln q(x)  - \ln p(x)] + \ln Z_{KL}
    \\
    Z_{KL} & = \mathbb{E}\Bigl[\frac{p(x)}{q(x)}\Bigr] \approx \frac{1}{S} \sum_{i=1}^S w_i
    \\
    \text{ESS} & = \frac{\sigma_{x \sim \mathcal{U}}^2[p(x)]}{\sigma^2\Bigl[\frac{p(x)}{q(x)}\Bigr]} \approx \frac{\Bigl(\sum_{i=1}^S w_i\Bigr)^2}{\sum_{i=1}^S w_i^2}
\end{align}

Figure \ref{fig:flows_fig}\textbf{A} shows a marginal density estimate of samples drawn from the \(p_\mathcal{F}^{mog}\) target density using our hit-and-run (HR) sampler. These samples serve as the training data for all models of this target density. In contrast, Figure \ref{fig:flows_fig}\textbf{C} displays samples from the uniform distribution \(p_\mathcal{F}^{\mathcal{U}}\), which is used as the base distribution for the Euclidean flow.

For the Aitchison flow, the polytope \(\mathcal{F}\) has \(V+1=14\) vertices.
To verify that the $\ilr$ transform correctly recovers the affine subspace corresponding to the image of the manifold \(\Lambda\), we compute the singular value decomposition (SVD) of Equation~\eqref{eq:ilr_svd} on 5000 $\ilr$ coordinates. The first five singular values, sorted by magnitude, are: 
\[
6.3\times10^{2}, \quad 2.7\times10^{2}, \quad 1.9\times10^{2}, \quad 6.3\times10^{1}, \quad 1.4\times10^{-13}.
\]
This rapid decay confirms that the effective dimension of the data is indeed \(K\), and that we can numerically recover the affine subspace for the image of \(\Lambda\).

For the \(p_\square^{mog}\) density, the training data is shown in Figure \ref{fig:Kflows}\textbf{A} and samples from the trained Euclidean flow are presented in panel \textbf{B}. A quantitative comparison between the models is provided in Table \ref{tab:model_comparison}.

\renewcommand{\arraystretch}{1.5}
\begin{table}[ht]
  \centering
  \begin{tabular}{lcccccc}
    \toprule
    Model & target &  base & dim & KL [nats] & ESS (\%) & outside (\%)\\
    \midrule
    \(q^{spline}\) & \(p^{mog}_\mathcal{F}\) & \(p^{\mathcal{U}}_\square\) & 4   & 6.747e-01  & 85.7 & -
    \\
    \(q^{eucl}\) & \(p^{mog}_\mathcal{F}\) & \(p^{\mathcal{U}}_\mathcal{F}\) & 4  & 8.440e-01 &  80.2 & 5.9
    \\
    \(q^{ball}\) & \(p^{mog}_\mathcal{F}\) & \(p^{\mathcal{U}}_\mathbb{B}\) & 4  & 8.985e-01  &  63.3 & -
    \\
    \(q^{ait}\) & \(p^{mog}_\mathcal{F}\) & \(p_{\mathbb{R}}^\mathcal{N}\) & 4  & 5.274e-01  &  79.3 & -
    \\
    \(q^{eucl}\) & \(p^{mog}_\square\) & \(p^{\mathcal{U}}_\square\) & 20  & 8.236e-02  &  19.8 & 12.5
    \\
    \bottomrule
    \vspace{2pt}
  \end{tabular}
  \caption{KL divergence (nats) w.r.t. target density and effective sample size (ESS) for 20k samples from every trained flow.}
  \label{tab:model_comparison}
\end{table}

We also compared the model architectures, hyper-parameters, as well as training and inference times for all models.
The results of this comparison are displayed in  Table \ref{tab:training_stats} in Appendix \ref{app:parameters}.
This Appendix elaborates on implementation details and is useful for practitioners that need to make choices based on their needs.

\newpage 

\section{Discussion}

The flows targeting \(p_\mathcal{F}^{mog}\) (see Table~\ref{tab:model_comparison}) exhibit similar performance, though key distinctions emerge. 
Riemannian flow matching ensures that no samples fall outside the polytope, whereas Euclidean flows occasionally produce out-of-polytope samples. 
In higher dimensions the proportion of volume near the facets increases, leading to more out-of-polytope samples for Euclidean CNFs. 
This is supported by both Figure~\ref{fig:Kflows} and Table~\ref{tab:model_comparison}.

For Riemannian CNFs operating directly on the convex polytope, higher dimensions would necessitate more frequent evaluations of the quadratic program projection of Equation~\eqref{eq:quad_proj_prog}, motivating the use of Riemannian ball flows. 
Although the ball flow avoids out-of-polytope samples, its lower ESS at comparable KL divergence indicates that the tail behavior may be less accurately captured.

In conclusion, modeling distributions on polytopes with Euclidean CNFs, especially in high dimensions, faces two challenges.
First, the base density is typically only approximately known; second, many samples produced by the flow may lie outside the polytope. 
Given the H-rperesentation of a polytope, these issues motivate transforming the polytope into a ball where the geometry is easier to manage.

When the ball serves as the target manifold, differences between CNFs and discrete flows become clear in terms of training and inference times (see Table~\ref{tab:training_stats}). 
Training a CNF via flow matching is nearly an order of magnitude faster than training a discrete flow, largely because the latter uses an autoregressive architecture. 
Although our analysis required both sample generation and the evaluation of the divergence integral for CNFs, which incurs additional overhead from automatic differentiation, a faster (albeit less accurate) alternative such as Hutchinson’s trace estimator \cite{grathwohl_ffjord_2018} could be employed.
Moreover, when only sampling is needed, the inference time difference between discrete flows and CNFs is negligible. 
Overall, the Riemannian CNF on the ball manifold offers significant training time advantages, with only a marginal increase in inference time.

We have also demonstrated that it is possible to construct flows for polytopes when only the V-representation is available. 
The original motivation for this construction was to handle high-dimensional polytopes where converting to the H-representation is computationally infeasible. 
Although our experiments were conducted in a \(K=4,\, V+1 = 14\) dimensional setting, where this motivation is less compelling, there remain important challenges. 
In higher dimensions, polytopes typically possess a vast number of vertices (e.g., a \(K=20\) dimensional hypercube has \(2^{20}=1048576\) vertices), causing the elements of \(\lambda^{\mec}\) to be very close to zero. 
This leads to numerical instabilities when solving the associated quadratic program, Equation~\eqref{eq:max_ent_coor}, to compute the $\mec$. 
Moreover, solving a \((V+1)\)-dimensional quadratic program for each data point can be computationally prohibitive, even though these computations can be parallelized. 
Only in very high dimensions would this approach potentially outperform converting to the H-representation and modeling flows that way. 
We thus consider the construction of Aitchison flows as an initial step toward modeling flows on polytopes in the V-representation.

\subsection{Related work \& outlook}

As detailed in Appendix \ref{appendix:multi_prop_hr_sampling}, extensive research over the past decades has focused on sampling from uniform densities over polytopes. 
In contrast, modeling non-uniform densities on these domains has primarily been explored within the framework of Bayesian inference for \ce{$^{13}$C}-MFA \cite{theorell_rethinking_2024, theorell_be_2017} and related flux applications \cite{heinonen_bayesian_2019}. 
These approaches typically rely on sampling-based methods, which are not amortized and require careful monitoring of convergence diagnostics. 
Here, we propose an alternative paradigm that leverages an amortized variational model for efficient density evaluation and sampling. 
Within the scope of \ce{$^{13}$C}-MFA, we envision this framework to not only accelerate posterior inference but also facilitate Bayesian optimal experiment design, i.e. choosing the isotopic labeling of substrates, by delivering fast and tractable density evaluations.

\subsection*{Acknowledgements}

Alexander Vitanov and Lorenzo Talamanca provided valuable feedback on the "Flows on barycentric coordinates" section and on the general style and readability of the paper.
This work was generously funded in part by the Swiss National Science Foundation (SNSF).
\newpage

\printbibliography

@article{heinonen_bayesian_2019,
	title = {Bayesian metabolic flux analysis reveals intracellular flux couplings},
	volume = {35},
	issn = {1367-4803},
	url = {https://doi.org/10.1093/bioinformatics/btz315},
	doi = {10.1093/bioinformatics/btz315},
	number = {14},
	urldate = {2025-03-07},
	journal = {Bioinformatics},
	author = {Heinonen, Markus and Osmala, Maria and Mannerström, Henrik and Wallenius, Janne and Kaski, Samuel and Rousu, Juho and Lähdesmäki, Harri},
	month = jul,
	year = {2019},
	pages = {i548--i557},
}

@incollection{john_extremum_2014,
	address = {Basel},
	title = {Extremum {Problems} with {Inequalities} as {Subsidiary} {Conditions}},
	isbn = {978-3-0348-0439-4},
	url = {https://doi.org/10.1007/978-3-0348-0439-4_9},
	language = {en},
	urldate = {2025-03-07},
	booktitle = {Traces and {Emergence} of {Nonlinear} {Programming}},
	publisher = {Springer},
	author = {John, Fritz},
	editor = {Giorgi, Giorgio and Kjeldsen, Tinne Hoff},
	year = {2014},
	doi = {10.1007/978-3-0348-0439-4_9},
	pages = {197--215},
}

@article{theorell_rethinking_2024,
	title = {Rethinking {13C}-metabolic flux analysis – {The} {Bayesian} way of flux inference},
	volume = {83},
	issn = {1096-7176},
	url = {https://www.sciencedirect.com/science/article/pii/S1096717624000508},
	doi = {10.1016/j.ymben.2024.03.005},
	urldate = {2025-03-07},
	journal = {Metabolic Engineering},
	author = {Theorell, Axel and Jadebeck, Johann F. and Wiechert, Wolfgang and McFadden, Johnjoe and Nöh, Katharina},
	month = may,
	year = {2024},
	pages = {137--149},
}

@article{theorell_be_2017,
	title = {To be certain about the uncertainty: {Bayesian} statistics for {13C} metabolic flux analysis},
	volume = {114},
	copyright = {© 2017 Wiley Periodicals, Inc.},
	issn = {1097-0290},
	shorttitle = {To be certain about the uncertainty},
	url = {https://onlinelibrary.wiley.com/doi/abs/10.1002/bit.26379},
	doi = {10.1002/bit.26379},
	language = {en},
	number = {11},
	urldate = {2025-03-07},
	journal = {Biotechnology and Bioengineering},
	author = {Theorell, Axel and Leweke, Samuel and Wiechert, Wolfgang and Nöh, Katharina},
	year = {2017},
	note = {\_eprint: https://onlinelibrary.wiley.com/doi/pdf/10.1002/bit.26379},
	pages = {2668--2684},
}

@article{wiechert_13c_2001,
	title = {{13C} {Metabolic} {Flux} {Analysis}},
	volume = {3},
	issn = {1096-7176},
	url = {https://www.sciencedirect.com/science/article/pii/S1096717601901879},
	doi = {10.1006/mben.2001.0187},
	number = {3},
	urldate = {2025-03-07},
	journal = {Metabolic Engineering},
	author = {Wiechert, Wolfgang},
	month = jul,
	year = {2001},
	pages = {195--206},
}

@article{antoniewicz_guide_2018,
	title = {A guide to {13C} metabolic flux analysis for the cancer biologist},
	volume = {50},
	copyright = {2018 The Author(s)},
	issn = {2092-6413},
	url = {https://www.nature.com/articles/s12276-018-0060-y},
	doi = {10.1038/s12276-018-0060-y},
	language = {en},
	number = {4},
	urldate = {2025-03-07},
	journal = {Experimental \& Molecular Medicine},
	author = {Antoniewicz, Maciek R.},
	month = apr,
	year = {2018},
	note = {Publisher: Nature Publishing Group},
	pages = {1--13},
}

@article{kumar_arviz_2019,
	title = {{ArviZ} a unified library for exploratory analysis of {Bayesian} models in {Python}},
	volume = {4},
	issn = {2475-9066},
	url = {https://joss.theoj.org/papers/10.21105/joss.01143},
	doi = {10.21105/joss.01143},
	language = {en},
	number = {33},
	urldate = {2025-03-05},
	journal = {Journal of Open Source Software},
	author = {Kumar, Ravin and Carroll, Colin and Hartikainen, Ari and Martin, Osvaldo},
	month = jan,
	year = {2019},
	pages = {1143},
}

@article{stimper_normflows_2023,
	title = {normflows: {A} {PyTorch} {Package} for {Normalizing} {Flows}},
	volume = {8},
	issn = {2475-9066},
	shorttitle = {normflows},
	url = {https://joss.theoj.org/papers/10.21105/joss.05361},
	doi = {10.21105/joss.05361},
	language = {en},
	number = {86},
	urldate = {2025-03-05},
	journal = {Journal of Open Source Software},
	author = {Stimper, Vincent and Liu, David and Campbell, Andrew and Berenz, Vincent and Ryll, Lukas and Schölkopf, Bernhard and Hernández-Lobato, José Miguel},
	month = jun,
	year = {2023},
	pages = {5361},
}

@article{barber_quickhull_1996,
	title = {The quickhull algorithm for convex hulls},
	volume = {22},
	issn = {0098-3500},
	url = {https://dl.acm.org/doi/10.1145/235815.235821},
	doi = {10.1145/235815.235821},
	number = {4},
	urldate = {2025-03-04},
	journal = {ACM Trans. Math. Softw.},
	author = {Barber, C. Bradford and Dobkin, David P. and Huhdanpaa, Hannu},
	month = dec,
	year = {1996},
	pages = {469--483},
}

@article{egozcue_isometric_2003,
	title = {Isometric {Logratio} {Transformations} for {Compositional} {Data} {Analysis}},
	volume = {35},
	issn = {1573-8868},
	url = {https://doi.org/10.1023/A:1023818214614},
	doi = {10.1023/A:1023818214614},
	language = {en},
	number = {3},
	urldate = {2025-02-28},
	journal = {Mathematical Geology},
	author = {Egozcue, J. J. and Pawlowsky-Glahn, V. and Mateu-Figueras, G. and Barceló-Vidal, C.},
	month = apr,
	year = {2003},
	pages = {279--300},
}

@misc{greenacre_aitchisons_2023,
	title = {Aitchison's {Compositional} {Data} {Analysis} 40 {Years} {On}: {A} {Reappraisal}},
	shorttitle = {Aitchison's {Compositional} {Data} {Analysis} 40 {Years} {On}},
	url = {http://arxiv.org/abs/2201.05197},
	doi = {10.48550/arXiv.2201.05197},
	urldate = {2025-02-28},
	publisher = {arXiv},
	author = {Greenacre, Michael and Grunsky, Eric and Bacon-Shone, John and Erb, Ionas and Quinn, Thomas},
	month = jan,
	year = {2023},
	note = {arXiv:2201.05197 [stat]},
}

@article{aitchison_statistical_1982,
	title = {The {Statistical} {Analysis} of {Compositional} {Data}},
	volume = {44},
	copyright = {© 1982 Royal Statistical Society},
	issn = {2517-6161},
	url = {https://onlinelibrary.wiley.com/doi/abs/10.1111/j.2517-6161.1982.tb01195.x},
	doi = {10.1111/j.2517-6161.1982.tb01195.x},
	language = {en},
	number = {2},
	urldate = {2025-02-28},
	journal = {Journal of the Royal Statistical Society: Series B (Methodological)},
	author = {Aitchison, J.},
	year = {1982},
	note = {\_eprint: https://onlinelibrary.wiley.com/doi/pdf/10.1111/j.2517-6161.1982.tb01195.x},
	pages = {139--160},
}

@incollection{lee_submersions_2012,
	address = {New York, NY},
	title = {Submersions, {Immersions}, and {Embeddings}},
	isbn = {978-1-4419-9982-5},
	url = {https://doi.org/10.1007/978-1-4419-9982-5_4},
	language = {en},
	urldate = {2025-02-28},
	booktitle = {Introduction to {Smooth} {Manifolds}},
	publisher = {Springer},
	author = {Lee, John M.},
	editor = {Lee, John M.},
	year = {2012},
	doi = {10.1007/978-1-4419-9982-5_4},
	pages = {77--97},
}

@article{avis_pivoting_1992,
	title = {A pivoting algorithm for convex hulls and vertex enumeration of arrangements and polyhedra},
	volume = {8},
	issn = {1432-0444},
	url = {https://doi.org/10.1007/BF02293050},
	doi = {10.1007/BF02293050},
	language = {en},
	number = {3},
	urldate = {2025-02-27},
	journal = {Discrete \& Computational Geometry},
	author = {Avis, David and Fukuda, Komei},
	month = aug,
	year = {1992},
	pages = {295--313},
}

@article{peskun_optimum_1973,
	title = {Optimum {Monte}-{Carlo} sampling using {Markov} chains},
	volume = {60},
	issn = {0006-3444},
	url = {https://doi.org/10.1093/biomet/60.3.607},
	doi = {10.1093/biomet/60.3.607},
	number = {3},
	urldate = {2025-02-24},
	journal = {Biometrika},
	author = {Peskun, P. H.},
	month = dec,
	year = {1973},
	pages = {607--612},
}

@article{metropolis_equation_1953,
	title = {Equation of {State} {Calculations} by {Fast} {Computing} {Machines}},
	volume = {21},
	issn = {0021-9606},
	url = {https://doi.org/10.1063/1.1699114},
	doi = {10.1063/1.1699114},
	number = {6},
	urldate = {2025-02-24},
	journal = {The Journal of Chemical Physics},
	author = {Metropolis, Nicholas and Rosenbluth, Arianna W. and Rosenbluth, Marshall N. and Teller, Augusta H. and Teller, Edward},
	month = jun,
	year = {1953},
	pages = {1087--1092},
}

@article{geyer_practical_1992,
	title = {Practical {Markov} {Chain} {Monte} {Carlo}},
	volume = {7},
	issn = {0883-4237},
	url = {https://www.jstor.org/stable/2246094},
	number = {4},
	urldate = {2025-02-24},
	journal = {Statistical Science},
	author = {Geyer, Charles J.},
	year = {1992},
	note = {Publisher: Institute of Mathematical Statistics},
	pages = {473--483},
}

@misc{sun_polytopewalk_2024,
	title = {{PolytopeWalk}: {Sparse} {MCMC} {Sampling} over {Polytopes}},
	shorttitle = {{PolytopeWalk}},
	url = {http://arxiv.org/abs/2412.06629},
	doi = {10.48550/arXiv.2412.06629},
	urldate = {2025-02-24},
	publisher = {arXiv},
	author = {Sun, Benny and Chen, Yuansi},
	month = dec,
	year = {2024},
	note = {arXiv:2412.06629 [stat]},
}

@article{antoniewicz_elementary_2007,
	title = {Elementary metabolite units ({EMU}): {A} novel framework for modeling isotopic distributions},
	volume = {9},
	issn = {1096-7176},
	shorttitle = {Elementary metabolite units ({EMU})},
	url = {https://www.sciencedirect.com/science/article/pii/S109671760600084X},
	doi = {10.1016/j.ymben.2006.09.001},
	number = {1},
	urldate = {2025-02-24},
	journal = {Metabolic Engineering},
	author = {Antoniewicz, Maciek R. and Kelleher, Joanne K. and Stephanopoulos, Gregory},
	month = jan,
	year = {2007},
	pages = {68--86},
}

@article{berbee_1987,
	title = {Hit-and-run algorithms for the identification of nonredundant linear inequalities},
	volume = {37},
	issn = {1436-4646},
	url = {https://doi.org/10.1007/BF02591694},
	doi = {10.1007/BF02591694},
	language = {en},
	number = {2},
	urldate = {2025-02-24},
	journal = {Mathematical Programming},
	author = {Berbee, H. C. P. and Boender, C. G. E. and Rinnooy Ran, A. H. G. and Scheffer, C. L. and Smith, R. L. and Telgen, J.},
	month = jun,
	year = {1987},
	pages = {184--207},
}

@article{lee_geodesic_2022,
	title = {Geodesic {Walks} in {Polytopes}},
	volume = {51},
	issn = {0097-5397},
	url = {https://epubs.siam.org/doi/abs/10.1137/17M1145999},
	doi = {10.1137/17M1145999},
	number = {2},
	urldate = {2025-02-24},
	journal = {SIAM Journal on Computing},
	author = {Lee, Yin Tat and Vempala, Santosh},
	month = apr,
	year = {2022},
	note = {Num Pages: STOC17-488
Publisher: Society for Industrial and Applied Mathematics},
	pages = {STOC17--400},
}

@techreport{tjelmeland_using_2004,
	title = {Using all {Metropolis}–{Hastings} proposals to estimate mean values},
	author = {Tjelmeland, Hakon},
	year = {2004},
}

@article{hormann_maximum_2008,
	title = {Maximum {Entropy} {Coordinates} for {Arbitrary} {Polytopes}},
	volume = {27},
	copyright = {© 2008 The Author(s) Journal compilation © 2008 The Eurographics Association and Blackwell Publishing Ltd.},
	issn = {1467-8659},
	url = {https://onlinelibrary.wiley.com/doi/abs/10.1111/j.1467-8659.2008.01292.x},
	doi = {10.1111/j.1467-8659.2008.01292.x},
	language = {en},
	number = {5},
	urldate = {2025-02-17},
	journal = {Computer Graphics Forum},
	author = {Hormann, K. and Sukumar, N.},
	year = {2008},
	note = {\_eprint: https://onlinelibrary.wiley.com/doi/pdf/10.1111/j.1467-8659.2008.01292.x},
	pages = {1513--1520},
}

@misc{davis_fisher_2024,
	title = {Fisher {Flow} {Matching} for {Generative} {Modeling} over {Discrete} {Data}},
	url = {http://arxiv.org/abs/2405.14664},
	doi = {10.48550/arXiv.2405.14664},
	urldate = {2025-02-14},
	publisher = {arXiv},
	author = {Davis, Oscar and Kessler, Samuel and Petrache, Mircea and Ceylan, İsmail İlkan and Bronstein, Michael and Bose, Avishek Joey},
	month = oct,
	year = {2024},
	note = {arXiv:2405.14664 [cs]},
}

@misc{lipman_flow_2024,
	title = {Flow {Matching} {Guide} and {Code}},
	url = {http://arxiv.org/abs/2412.06264},
	doi = {10.48550/arXiv.2412.06264},
	urldate = {2025-02-10},
	publisher = {arXiv},
	author = {Lipman, Yaron and Havasi, Marton and Holderrieth, Peter and Shaul, Neta and Le, Matt and Karrer, Brian and Chen, Ricky T. Q. and Lopez-Paz, David and Ben-Hamu, Heli and Gat, Itai},
	month = dec,
	year = {2024},
	note = {arXiv:2412.06264 [cs]},
}

@misc{chen_neural_2019,
	title = {Neural {Ordinary} {Differential} {Equations}},
	url = {http://arxiv.org/abs/1806.07366},
	doi = {10.48550/arXiv.1806.07366},
	urldate = {2025-02-10},
	publisher = {arXiv},
	author = {Chen, Ricky T. Q. and Rubanova, Yulia and Bettencourt, Jesse and Duvenaud, David},
	month = dec,
	year = {2019},
	note = {arXiv:1806.07366 [cs]},
}

@misc{grathwohl_ffjord_2018,
	title = {{FFJORD}: {Free}-form {Continuous} {Dynamics} for {Scalable} {Reversible} {Generative} {Models}},
	shorttitle = {{FFJORD}},
	url = {http://arxiv.org/abs/1810.01367},
	doi = {10.48550/arXiv.1810.01367},
	urldate = {2025-02-10},
	publisher = {arXiv},
	author = {Grathwohl, Will and Chen, Ricky T. Q. and Bettencourt, Jesse and Sutskever, Ilya and Duvenaud, David},
	month = oct,
	year = {2018},
	note = {arXiv:1810.01367 [cs]},
}

@misc{chen_flow_2024,
	title = {Flow {Matching} on {General} {Geometries}},
	url = {http://arxiv.org/abs/2302.03660},
	doi = {10.48550/arXiv.2302.03660},
	urldate = {2025-02-10},
	publisher = {arXiv},
	author = {Chen, Ricky T. Q. and Lipman, Yaron},
	month = feb,
	year = {2024},
	note = {arXiv:2302.03660 [cs]},
}

@misc{lipman_flow_2023,
	title = {Flow {Matching} for {Generative} {Modeling}},
	url = {http://arxiv.org/abs/2210.02747},
	doi = {10.48550/arXiv.2210.02747},
	urldate = {2025-02-10},
	publisher = {arXiv},
	author = {Lipman, Yaron and Chen, Ricky T. Q. and Ben-Hamu, Heli and Nickel, Maximilian and Le, Matt},
	month = feb,
	year = {2023},
	note = {arXiv:2210.02747 [cs]},
}

@article{smith_efficient_1984,
	title = {Efficient {Monte} {Carlo} {Procedures} for {Generating} {Points} {Uniformly} {Distributed} over {Bounded} {Regions}},
	volume = {32},
	issn = {0030-364X},
	url = {https://pubsonline.informs.org/doi/10.1287/opre.32.6.1296},
	doi = {10.1287/opre.32.6.1296},
	number = {6},
	urldate = {2025-02-10},
	journal = {Operations Research},
	author = {Smith, Robert L.},
	month = dec,
	year = {1984},
	note = {Publisher: INFORMS},
	pages = {1296--1308},
}

@article{kaufman_direction_1998,
	title = {Direction {Choice} for {Accelerated} {Convergence} in {Hit}-and-{Run} {Sampling}},
	volume = {46},
	issn = {0030-364X},
	url = {https://pubsonline.informs.org/doi/10.1287/opre.46.1.84},
	doi = {10.1287/opre.46.1.84},
	number = {1},
	urldate = {2025-02-10},
	journal = {Operations Research},
	author = {Kaufman, David E. and Smith, Robert L.},
	month = feb,
	year = {1998},
	note = {Publisher: INFORMS},
	pages = {84--95},
}

@misc{mathieu_riemannian_2020,
	title = {Riemannian {Continuous} {Normalizing} {Flows}},
	url = {http://arxiv.org/abs/2006.10605},
	doi = {10.48550/arXiv.2006.10605},
	urldate = {2025-02-07},
	publisher = {arXiv},
	author = {Mathieu, Emile and Nickel, Maximilian},
	month = dec,
	year = {2020},
	note = {arXiv:2006.10605 [stat]},
}

@article{haraldsdottir_chrr_2017,
	title = {{CHRR}: coordinate hit-and-run with rounding for uniform sampling of constraint-based models},
	volume = {33},
	issn = {1367-4803},
	shorttitle = {{CHRR}},
	url = {https://doi.org/10.1093/bioinformatics/btx052},
	doi = {10.1093/bioinformatics/btx052},
	number = {11},
	urldate = {2025-02-06},
	journal = {Bioinformatics},
	author = {Haraldsdóttir, Hulda S and Cousins, Ben and Thiele, Ines and Fleming, Ronan M.T and Vempala, Santosh},
	month = jun,
	year = {2017},
	pages = {1741--1743},
}

@article{quek_openflux_2009,
	title = {{OpenFLUX}: efficient modelling software for {13C}-based metabolic flux analysis},
	volume = {8},
	issn = {1475-2859},
	shorttitle = {{OpenFLUX}},
	url = {https://doi.org/10.1186/1475-2859-8-25},
	doi = {10.1186/1475-2859-8-25},
	number = {1},
	urldate = {2025-02-06},
	journal = {Microbial Cell Factories},
	author = {Quek, Lake-Ee and Wittmann, Christoph and Nielsen, Lars K. and Krömer, Jens O.},
	month = may,
	year = {2009},
	pages = {25},
}

@misc{durkan_cubic-spline_2019,
	title = {Cubic-{Spline} {Flows}},
	url = {http://arxiv.org/abs/1906.02145},
	doi = {10.48550/arXiv.1906.02145},
	urldate = {2025-01-29},
	publisher = {arXiv},
	author = {Durkan, Conor and Bekasov, Artur and Murray, Iain and Papamakarios, George},
	month = jun,
	year = {2019},
	note = {arXiv:1906.02145 [stat]},
}

@misc{durkan_neural_2019,
	title = {Neural {Spline} {Flows}},
	url = {http://arxiv.org/abs/1906.04032},
	doi = {10.48550/arXiv.1906.04032},
	urldate = {2025-01-29},
	publisher = {arXiv},
	author = {Durkan, Conor and Bekasov, Artur and Murray, Iain and Papamakarios, George},
	month = dec,
	year = {2019},
	note = {arXiv:1906.04032 [stat]},
}

@article{zhang_numerical_2003,
	title = {On {Numerical} {Solution} of the {Maximum} {Volume} {Ellipsoid} {Problem}},
	volume = {14},
	issn = {1052-6234},
	url = {https://epubs.siam.org/doi/10.1137/S1052623401397230},
	doi = {10.1137/S1052623401397230},
	number = {1},
	urldate = {2025-01-25},
	journal = {SIAM Journal on Optimization},
	author = {Zhang, Yin and Gao, Liyan},
	month = jan,
	year = {2003},
	note = {Publisher: Society for Industrial and Applied Mathematics},
	pages = {53--76},
}

@article{theorell_polyround_2022,
	title = {{PolyRound}: polytope rounding for random sampling in metabolic networks},
	volume = {38},
	issn = {1367-4803},
	shorttitle = {{PolyRound}},
	url = {https://doi.org/10.1093/bioinformatics/btab552},
	doi = {10.1093/bioinformatics/btab552},
	number = {2},
	urldate = {2025-01-25},
	journal = {Bioinformatics},
	author = {Theorell, Axel and Jadebeck, Johann F and Nöh, Katharina and Stelling, Jörg},
	month = jan,
	year = {2022},
	pages = {566--567},
}

@phdthesis{liphardt_efficient_2018,
	type = {Doctoral {Thesis}},
	title = {Efficient computational methods for sampling-based metabolic flux analysis},
	copyright = {http://rightsstatements.org/page/InC-NC/1.0/},
	url = {https://www.research-collection.ethz.ch/handle/20.500.11850/271574},
	language = {en},
	urldate = {2025-01-25},
	school = {ETH Zurich},
	author = {Liphardt, Thomas},
	year = {2018},
	doi = {10.3929/ethz-b-000271574},
	note = {Accepted: 2018-06-22T10:34:22Z},
}

@misc{rezende_normalizing_2020,
	title = {Normalizing {Flows} on {Tori} and {Spheres}},
	url = {http://arxiv.org/abs/2002.02428},
	doi = {10.48550/arXiv.2002.02428},
	urldate = {2025-01-22},
	publisher = {arXiv},
	author = {Rezende, Danilo Jimenez and Papamakarios, George and Racanière, Sébastien and Albergo, Michael S. and Kanwar, Gurtej and Shanahan, Phiala E. and Cranmer, Kyle},
	month = jul,
	year = {2020},
	note = {arXiv:2002.02428 [stat]},
}

\appendix

\section{Multi-proposal hit-and-run sampling of arbitrary densities over polytopes}
\label{appendix:multi_prop_hr_sampling}

Uniform sampling from polytopes is a topic that has been widely studied for decades; see for instance \cite{berbee_1987, lee_geodesic_2022, sun_polytopewalk_2024}. For applications such as \ce{$^{13}$C}-MFA, we are often interested in sampling non-uniform densities over a polytope. 
For instance, density \(\pi\) could represent the posterior over fluxes, where at every proposal a labeling state needs to be simulated in order to compute the likelihood. 
The Elementary Metabolic Unit (EMU) algorithm \cite{antoniewicz_elementary_2007} is one example of such a simulation algorithm. 
Labeling simulations generally consist of solving a cascade of linear systems, which can easily be parallelized on a GPU.

The number of density evaluations at every step of a sampling algorithm is \(L \times M\), where \(L\) is the number of chains and \(M\) is the number of proposals (typically \(M=1\)). 
Markov chain approaches are inherently sequential, and if the stationary distribution is complex, running more chains to increase parallelism might not speed things up, since individual chains need to converge \cite{geyer_practical_1992}. 
For this reason, we developed the multi-proposal hit-and-run algorithm (Algorithm~\ref{algo:mcmc}), where parallelism is increased by evaluating the density of multiple proposals (\(M>1\)). 
In this publication, we are not dealing with labeling simulations for density evaluation but instead try to sample from a mixture of Gaussians constrained to a polytope. 
In this case too, our algorithm is a sensible choice since it allows for a tunable proposal distribution, the choice of which can significantly influence the convergence of the Markov chains.

\begin{figure}[h]
\begin{algorithm}[H]
\caption{Multiple proposal Hit-and-Run sampling of distributions with polytope support}\label{algo:mcmc}
\DontPrintSemicolon
\KwIn{\(\pmb{A}, b\) defining a full-dimensional polytope \(\mathcal{F} = \{v \in  \mathbb{R}^{K} \mid \pmb{A} \cdot v \leq b \}\)} 
\KwIn{\(N\) number of samples in a single chain} 
\KwIn{\(M\) number of proposals to evaluate in a single step of the chain} 
\KwIn{\(\pi\) target density}
\KwIn{\(\mathscr{q}\) proposal density} 
\KwOut{\(\mathcal{Y}\) samples from (approx.) posterior}
\nonl\;
\SetKwProg{Fn}{function}{:}{end function}
\Fn{\textnormal{chord\_extremes}(v,s,\(\pmb{A}, b\))}{
  \(d^s = \pmb{A} \cdot s\)\;
  \(d^v = b - \pmb{A} \cdot v\)\;
  \(\alpha = d^v \oslash d^s\)\;
  \(\alpha^{min} = \max(\alpha \mid \alpha \leq 0)\)\;
  \(\alpha^{max} = \min(\alpha \mid \alpha \geq 0)\)\;
  \Return{\(\alpha^{min}, \alpha^{max}\)}
}
\SetKwProg{Fn}{function}{:}{end function}
\Fn{\textnormal{MCMC}(\(\pmb{A}, b, N, M, \pi, \mathscr{q}\))}{
  Sample initial point from ball: \(v^0 \sim \mathcal{U}(\mathbb{B}^K)\)\;
  \(\mathcal{Y} \gets \{v^0\}\)\;
  \(i \gets 0\)\;
  \While{\(i < N\)}{
    Sample direction from sphere: \(s \sim \mathcal{U}(\mathbb{S}^{K-1})\)\;
    \(\alpha^{min}, \alpha^{max} \gets \text{chord\_extremes}(v^0, s, \pmb{A}, b)\)\;
    \(\alpha \gets \Bigl[\alpha_i \sim \mathscr{q}(\alpha ; \alpha^{min}, \alpha^{max}) \quad \forall \, i \in \{1,\ldots,M\}\Bigr]^T\)\;
    Proposals on the chord: \(\mathcal{C} \gets \{v^0\} \cup \{v^0 + s \cdot \alpha_i\}\)\;
    Compute weights \(w\) from Equation \eqref{eq:peskun} or \eqref{eq:barker} with proposals \(\mathcal{C}\)\;
    Accept proposal: \(k \sim \text{Categorical}(w)\)\;
    \(v^0 \gets \mathcal{C}_k\), \(\mathcal{Y} \gets \mathcal{Y} \cup \{\mathcal{C}_k\}\), \(i \gets i+1\)\;
  }
  \Return{\(\mathcal{Y}\)}
}
\end{algorithm}
\end{figure}

We base our multi-proposal MCMC on \textcite{tjelmeland_using_2004}. 
When evaluating multiple proposals, there generally are two choices for transition kernels whose stationary distribution is density \(\pi\). 
The first is the Barker transition kernel of Equation \eqref{eq:barker}, which is the one used in the original \(M=1\) Metropolis-Hastings algorithm \cite{metropolis_equation_1953}. 
It is rarely seen in practice anymore, since \textcite{peskun_optimum_1973} proved that the asymptotic variance of estimators is lower when using the transition kernel of Equation \eqref{eq:peskun}.

\begin{align}
    \mathscr{q}(v^i \mid v^{/i}) & = \prod_{j \in \{0:M\}, j\neq i} \mathscr{q}(v^i \mid v^j) & \text{independent proposal probabilities} 
    \\
    &= \prod_{j \in \{0:M\}, j\neq i} \mathscr{q}(\alpha_i \mid \alpha_j) & \text{proposals on the chord}\label{eq:alpha_proposals}
    \\
    & \begin{cases}
    w_i (v^i) &= \frac{1}{M} \min \Bigl( 1,  \frac{\pi(v^i) \cdot \mathscr{q}(v^i \mid v^{/i})}{\pi(v^0) \cdot \mathscr{q}(v^0 \mid v^{/0})} \Bigr) \quad \forall i \in \{1,\ldots,M\}
    \\
    w_0 (v^0) &= 1 - \sum_{i \in \{1,\ldots,M\}}w_i
    \end{cases} & \text{Peskun transition weights} \label{eq:peskun}
    \\
    w_i & = \frac{\pi(v^i) \cdot \mathscr{q}(v^i \mid v^{/i})}{\sum_{i \in \{0:M\}}\pi(v^i) \cdot \mathscr{q}(v^i \mid v^{/i})} & \text{Barker transition weights} \label{eq:barker}
\end{align}

Note that for Algorithm~\ref{algo:mcmc} we do not sample proposals fully independently. 
Independent sampling would entail sampling a direction \(s\) for every proposal, but this would increase code complexity since the computation of the terms in Equation \eqref{eq:alpha_proposals} would become more cumbersome. 
Also note that the proposal distribution is defined over scalar values \(\alpha\). The two choices for proposal distribution are uniform: \(\mathscr{q} = \mathcal{U}(\alpha ; \alpha^{\min}, \alpha^{\max})\) and truncated normal: \(\mathscr{q} = \mathcal{N}(\alpha; \alpha^{\min}, \alpha^{\max}, \mu=0, \sigma^2)\). 
The truncated normal is centered on the current point (hence \(\mu = 0\)) and has a tunable parameter \(\sigma^2\). Our algorithm allows for further tuning through the specification of a covariance matrix \(\pmb{\Sigma}\in \mathbb{R}^{K\times K}\). 
The variance along a chord can then be computed as follows: \(\sigma^2 = s^T \cdot \pmb{\Sigma} \cdot s\).

For the \(p_\mathcal{F}^{mog}\) target density, we used a uniform proposal density with 3 proposals per step (excluding the current state) and adopted the Peskun transition kernel (Equation \eqref{eq:peskun}). 
We ran 8 chains in parallel, discarding the first 1000 steps as burn-in and then thinning the chains by recording every 15th sample.

For the \(p_\mathcal{F}^{unif}\) density, we configured the sampler with a uniform proposal density that generates a single proposal per step, combined with a Peskun transition kernel. 
This setup corresponds to the classical Metropolis-Hastings algorithm. 
Again, we ran 8 chains in parallel, with a 1000-step burn-in, and in this case every 10th sample was retained.

Convergence metrics for these samplings, including the \(\hat{R}\) statistic and effective sample size (ESS), are summarized in Table~\ref{tab:convergence_stats}. 
These metrics were computed using the \verb|arviz| package \cite{kumar_arviz_2019}. 
We do not report the convergence statistics for the 20-dimensional \(p_\square^{mog}\) density here, but these details are available in the accompanying Jupyter notebooks.

\begin{table}[ht]
\centering
\begin{tabular}{llllll}
\toprule
Target          & Statistic  & R\_v7 &	R\_f\_out &	R\_biomass &	R\_h\_out \\
\midrule
\multirow{2}{*}{\(p_\mathcal{F}^{mog}\), \(S=105k\)}  & ESS (\%) & 19.9 &	56.3 &	16.6 &	14.7 \\
                            & \(\hat{R}\)       & 1.000312 &	1.000115 &	1.000488 &	1.000358 \\
                            \hline
\multirow{2}{*}{\(p_\mathcal{F}^{unif}\), \(S=125k\)} & ESS (\%) & 60.8 &	58.6 &	80.6 &	61.3 \\
                            & \(\hat{R}\)       & 1.000033 &	0.999992 &	1.000044 &	1.000013 \\
\end{tabular}
\caption{Convergence metrics for the MCMC sampling of a 4-dimensional polytope. ESS = effective sample size.}\label{tab:convergence_stats}
\end{table}

\section{Model parameters}
\label{app:parameters}

In this Section, we review the model architectures along with the training and inference performance of the flows employed in our experiments. 
The results are summarized in Table \ref{tab:training_stats}. 
The \textit{div} column indicates the time required to generate 20k samples from the flow, including the computation of the log determinant or log divergence integral.
Because CNFs incur additional overhead for evaluating the divergence integral, we also report the pure sampling time in the \textit{sample} column.

\renewcommand{\arraystretch}{1.5}
\begin{table}[ht]
  \centering
  \begin{tabular}{lcccccccccc}
    \toprule
    Model & target &  dim & hid. lay. & hid. dim. & lr & epochs & batch size & train (s) & div (s) & sample (s)\\
    \midrule
    \(q^{spline}\) & \(p^{mog}_\mathcal{F}\) &  4 & 4  & 64 & 4e-3 & 35 & 12288 & 1504 & 1.5 & -
    \\
    \(q^{eucl}\) & \(p^{mog}_\mathcal{F}\) &  4  & 6 & 512 & 1e-3 & 35 & 8192 & 200 & 74 & 3.8
    \\
    \(q^{ball}\) & \(p^{mog}_\mathcal{F}\) &  4  &  6 &  512 & 1e-3 &  35 & 8192 & 247 & 32 & 1.6 
    \\
    \(q^{ait}\) & \(p^{mog}_\mathcal{F}\) &  4  & 6  &  512 & 1e-3 & 50 & 8192 & 213 & 12 & 1.7
    \\
    \(q^{eucl}\) & \(p^{mog}_\square\) &  20  & 6  &  1024 & 1e-3 & 35 & 8192 & 302 & 998 & 11.4
    \\
    \bottomrule
    \vspace{2pt}
  \end{tabular}
  \caption{Comparison of key architectural parameters and corresponding training and inference times for various models. Inference times are separated into divergence (for continuous flows) or determinant (for discrete flows) computation in the \textit{div} column and only sampling times in the \textit{sample} column.}
  \label{tab:training_stats}
\end{table}

For the \(q^{spline}\) model, note that the \textit{hid. dim.} is noted per transformation. 
For \(q^{spline}\), the following architecture was used. We used a flow of 10 transformations interspersed with permutation layers. 
Each transformation is an auto-regressive rational quadratic spline flow \cite{durkan_neural_2019} where the \(\theta\) dimension was modeled as a circular spline flow \cite{rezende_normalizing_2020}. Each spline had 30 bins (i.e., 31 knots, including end-points). 
We adapted the flows presented in the \verb|normflows| package \cite{stimper_normflows_2023}. 
When sampling from \(q^{spline}\), both the samples and log determinant are returned in one go, hence there being no value in the \textit{sample} column. 
Although we did not perform a systematic hyperparameter search, our experiments indicate that reducing the complexity of the model (fewer hidden layers, fewer transformations, or lower hidden dimensions) degrades performance.

For inference with all CNFs, we used a midpoint numerical integrator with a step size of 0.05. 
In particular, for \(q^{ball}\) we utilized the Riemannian version of the midpoint solver. 
Our CNF implementations and training procedures are based on adaptations of the algorithms presented in the \verb|flow_matching| package \cite{lipman_flow_2024}. 
We computed the divergence integral using automatic differentiation to ensure accurate estimates, and we implemented the forward divergence (integrating from \(t=0\) to \(t=1\)) estimation ourselves, as this functionality was not yet available in the package.

All experiments were performed on a laptop with an Intel(R) Core(TM) i7-7700HQ @ 2.80GHz CPU and an NVIDIA GeForce GTX 1060 6GB GPU, which is CUDA enabled and was utilized for all experiments.
\end{document}